\documentclass{article}
\usepackage[accepted]{icml2025}

\usepackage{graphicx} 
\usepackage[utf8]{inputenc}
\usepackage{kotex} 
\usepackage{multido}
\usepackage{mathdots}
\usepackage{mathtools}
\usepackage{amsmath}
\usepackage{graphicx, animate}
\usepackage{ragged2e}
\usepackage{paracol}
\usepackage{tcolorbox}
\usepackage{array}
\usepackage{adjustbox}
\usepackage{hyperref}
\usepackage{caption}
\usepackage{mathrsfs}
\usepackage{amssymb}
\usepackage{natbib}
\usepackage{booktabs}
\usepackage{multirow}
\graphicspath{{figures/}}
\usepackage{abbreviations}
\usepackage[T1]{fontenc}
\usepackage[capitalize,noabbrev]{cleveref}
\usepackage{natbib}
\usepackage{threeparttable}
\usepackage{algorithm}  
\usepackage{bbm} 
\usepackage{dsfont}
\usepackage{subcaption}
\usepackage{enumitem}
\usepackage{cuted}

\DeclareRobustCommand{\bbone}{\text{\usefont{U}{bbold}{m}{n}1}}
\icmltitlerunning{Optimal and Practical Batched Linear Bandit Algorithm}

\begin{document}
\twocolumn[
\icmltitle{
Optimal and Practical Batched Linear Bandit Algorithm
}

\icmlsetsymbol{equal}{*}

\begin{icmlauthorlist}
\icmlauthor{Sanghoon Yu}{snu}
\icmlauthor{Min-hwan Oh}{snu}
\end{icmlauthorlist}

\icmlaffiliation{snu}{Seoul National University, Seoul, Republic of Korea}

\icmlcorrespondingauthor{Min-hwan Oh}{minoh@snu.ac.kr}

\icmlkeywords{Bandit algorithms, Batched linear bandit, Optimal design}

\vskip 0.3in
]
\printAffiliationsAndNotice{}
\begin{abstract}
We study the linear bandit problem under limited adaptivity, known as the batched linear bandit. While existing approaches can achieve near-optimal regret in theory, they are often computationally prohibitive or underperform in practice. We propose \texttt{BLAE}, a novel batched algorithm that integrates arm elimination with regularized G-optimal design, achieving the minimax optimal regret (up to logarithmic factors in $T$) in both large-$K$ and small-$K$ regimes for the first time, while using only $\Ocal(\log\log T)$ batches. Our analysis introduces new techniques for batch-wise optimal design and refined concentration bounds. Crucially, \texttt{BLAE} demonstrates low computational overhead and strong empirical performance, outperforming state-of-the-art methods in extensive numerical evaluations. Thus, \texttt{BLAE} is the first algorithm to combine provable minimax-optimality in all regimes and practical superiority in batched linear bandits.
\end{abstract}

\section{Introduction} \label{sec:introduction}

Stochastic linear bandits form a cornerstone of sequential decision-making and online learning, enabling an agent to repeatedly select actions from a feature-based action set and observe stochastic rewards drawn from an unknown linear model~\citep{abe1999associative, auer2002using, dani2008stochastic, li2010contextual, abbasi2011improved, lattimore2020bandit}.
This framework elegantly captures settings where the action space can be large but structurally low-dimensional, as actions naturally embed into a feature space.
Such structure enables efficient generalization and has been successfully applied across a wide range of domains, including  clinical trials and precision medicine~\citep{lu2021bandit} to recommender systems~\cite{li2010contextual} and  inventory control~\citep{jin2021shrinking}.

Despite the rich literature on \textit{fully adaptive} linear bandit algorithms—those that update policies or parameters every round~\citep{auer2002using,abe2003reinforcement,dani2008stochastic,chu2011contextual,abbasi2011improved,li2019nearly,kirschner2021asymptotically}, many practical systems often cannot afford such frequent updates because of computational or operational costs.
Large-scale recommendation systems, for example, cannot retrain policies every interaction, and in clinical trials, continually changing treatments can be both logistically and ethically prohibitive. 
These constraints have motivated a growing body of work on \emph{batched} linear bandits, where policy updates 
occur in a small number of discrete \emph{batches}~\citep{abbasi2011improved,han2020sequential,esfandiari2021regret,ruan2021linear,hanna2023contexts,ren2024optimal,zhang2025almost}.


Batched linear bandit algorithms must strike a delicate balance between exploration and exploitation within each batch, aiming to minimize both the number of policy updates and the cumulative regret. Although recent work provides strong theoretical guarantees \citep{ren2024optimal,zhang2025almost}, their empiricial limitations may undermine the intended benefits of batched learning. 
In numerical experiments (see \cref{sec:experiment}), the method of \citet{ren2024optimal} tends to eliminate arms prematurely (often including the optimal arm) and \emph{suffers in practical performance}, while other methods update parameters rather too frequently~\citep{abbasi2011improved,esfandiari2021regret}.\footnote{Note that \citet[Section 5.1]{abbasi2011improved} introduced the rare-switching OFUL that attains a batch complexity of $\Ocal(d\log T)$.}  
Still others require such substantial computation that they become infeasible even at moderate problem scales~\citep{ruan2021linear,hanna2023contexts,zhang2025almost} (see \cref{sec:additional-experiments}). Much of this empirical inefficiency observed with the methods in \citet{ruan2021linear,hanna2023contexts,zhang2025almost} stems from algorithmic components tailored to contextual settings with time-varying features; when these algorithms are applied to the plain, fixed-feature linear bandit, they cannot exploit the static structure and therefore still incur excessive computation.

Consequently, despite recent progress, it remains unclear whether a batched linear bandit algorithm can simultaneously attain minimax-optimal regret and exhibit strong empirical performance. 
This open challenge motivates our central research question:

\begin{center}
    \emph{Can we design a provably optimal and practically superior batched linear bandit algorithm?}
\end{center}

A positive answer to the question above would bridge the gap between theory and practice, offering a concrete blueprint for algorithms that are both statistically sound and computationally viable for batched linear bandits.

\paragraph{Our Contributions.}
We present a practical and minimax-optimal batched linear bandit algorithm, \emph{Batched Linear bandit Algorithm with Elimination} (\texttt{BLAE}), that simultaneously addresses both computational efficiency and theoretical guarantees with practicality. We summarize our contributions below:

\begin{itemize}
\item \textbf{Novel Batched Algorithm and Regret Guarantees.}
We propose \texttt{BLAE}, which constructs a judiciously designed arm-elimination strategy integrated with regularized G-optimal design. BLAE achieves a worst-case regret  (\cref{thm:regret of algorithm}) of
\[
\Ocal\Bigl(\sqrt{dT}\,\bigl(\sqrt{\log(KT)} \wedge \sqrt{d + \log T}\bigr)\log\log T \Bigr),
\]
where $K$ is the number of arms, $d$ is the feature dimension, and $T$ is the total time horizon.
To the best of our knowledge, this is the tightest known bound for  batched linear bandits.
More importantly, \texttt{BLAE} is the first batched linear bandit algorithm that (up to logarithmic factors in $T$) simultaneously matches the lower bounds of $\Omega(d\sqrt{T})$ for the large-$K$ regime ($K \geq \Omega(e^d)$) and $\Omega\big(\sqrt{dT \log K}\big)$ for the small-$K$ regime ($K \leq \Ocal(e^d)$). 

\item \textbf{Batch Complexity.}
Notably, \texttt{BLAE} achieves the minimax optimal regret for both the large-$K$ and small-$K$ regimes
with minimal batch complexity of $\Ocal(\log\log T)$.

\item \textbf{Refined Technical Analysis.}
Key to our results are new analysis techniques for regularized G-optimal design under batch updates (\cref{lemma:regularized-G-optimal-design}), along with a concentration bounds that accommodate both batching and regularization (Lemmas \ref{lemma:concentration} and \ref{lemma:good-event}). 
Additionally, our refined analysis of efficient coverings of the unit sphere provides both theoretical guarantees and practical improvements (\cref{lemma 8}). 
These technical tools may be of independent interest for both batched and non-batched bandit research.

\item \textbf{Practical Superiority.}
Extensive numerical evaluations demonstrate that
\texttt{BLAE}
consistently and substantially outperforms existing batched linear bandit algorithms, offering a combination of provable near-optimality across all regimes and strong empirical performance.

\end{itemize}


Overall, \texttt{BLAE} demonstrates that provably optimal worst-case regret can be achieved alongside low batch complexity and tractable computation---key objectives in batched bandit research.
By uniting strong theoretical guarantees with robust empirical performance, \texttt{BLAE} helps resolve the longstanding tension between provable efficiency and practical viability in the batched linear bandit setting.

\begin{table*}[ht]
\centering
\begin{threeparttable}         
\caption{Worst-case regret and batch complexity comparison in batched linear bandits. \texttt{BLAE} achieves a regret bound of $\widetilde{\Ocal}\bigl(\sqrt{dT\log K}\wedge d\sqrt{T}\bigr)$, matching the best-known (and indeed minimax-optimal) results for fully adaptive linear bandit problems. 
Moreover, \texttt{BLAE} operates with only $\Ocal(\log\log T)$ batches, comparable to the lowest batch complexity among existing methods. The regret bound of \texttt{BLAE} simultaneously covers both the large-$K$ regime ($K \geq \Omega(e^d)$) with $\Ocal(d\sqrt{T})$ regret and the small-$K$ regime ($K \leq \Ocal(e^d)$) with $\Ocal(\sqrt{dT\log K})$ regret matching the lower bounds for the both regimes, thereby establishing the tightest known performance guarantees for batched linear bandits.}
\label{table:comparison}
\begin{tabular}{lll}
\toprule
\textbf{Paper} & \textbf{Worst-Case Regret} & \textbf{Batch Complexity} \\
\midrule
{\citet{abbasi2011improved}} & $\Ocal(d\sqrt{T}\log T)$ & $\Ocal(d\log T)$ \\
{\citet{esfandiari2021regret}} & $\Ocal(\sqrt{dT\log(KT)})$ & $\Ocal(\log T)$ 
\\
{\citet{ruan2021linear}} & $\Ocal(\sqrt{dT\log(dKT)\log d}\log\log T)$ & $\Ocal(\log \log T)$ 
\\
{\citet{hanna2023contexts}} & $\Ocal(d\sqrt{T\log T}\log\log T)$ & $\Ocal(\log \log T)$ \\
{\citet{ren2024optimal}} & $\Ocal(\sqrt{dT\log (KT)}\log\log T)$ & $\Ocal(\log \log T)$
\\
{\citet{zhang2025almost}} & $\Ocal(\sqrt{dT\log (dKT)\log T}\,\log(dT) \log\log T)$ & $\Ocal(\log \log T)$ 
\\
\midrule
{\citet{dani2008stochastic} (Lower bound)} & $\Omega(d\sqrt{T})$ & {--} \\
{\citet{zhou2019lecture14} (Lower bound)} & $\Omega(\sqrt{dT\log K})$ & {--} \\
\midrule
\textbf{This work} & \(\Ocal\big( \sqrt{dT}\big(\sqrt{\log(KT)}\wedge \sqrt{d+\log T}\big)\log\log T \big)\) & $\Ocal(\log \log T)$ \\
\bottomrule
\end{tabular}
\end{threeparttable}
\end{table*}

\section{Related Work} \label{sec:related work}

A growing body of work has investigated \emph{batched bandits}, in which the learner is allowed to update its policy only at a small number of discrete batch points, thereby reconciling near-optimal regret guarantees with a strict limit on adaptivity~\citep{abbasi2011improved,perchet2016batched,gao2019batched,han2020sequential,ruan2021linear,jin2021almost,jin2021double,hanna2023contexts,esfandiari2021regret,jin2024optimal,ren2024optimal,zhang2025almost}.

Several of the earlier contributions focus on the multi-armed setting~\citep{perchet2016batched,gao2019batched,jin2021almost,jin2021double}. Subsequent work extends batching to linear contextual bandits, beginning with Gaussian-type feature models~\citep{han2020sequential} and later generalized to adversarial features~\citep{esfandiari2021regret}. 
More recent algorithms attain near-optimal regret while reducing the batch complexity to the minimal $\mathcal{O}(\log\log T)$~\citep{ruan2021linear,hanna2023contexts,ren2024optimal,zhang2025almost}.

Despite these theoretical advances, many existing methods face practical obstacles.  Algorithms that rely on $\mathcal{O}(\log T)$ batches~\citep{abbasi2011improved,esfandiari2021regret} still update more frequently than the desirable $\mathcal{O}(\log\log T)$, increasing computational overhead. 
Approaches that do match the lower batch complexity often employ two-phase updates that split each batch into separate ``estimation'' and ``policy-learning'' stages~\citep{ruan2021linear,zhang2025almost}; this structure inflates runtime and may leave the information matrix ill-conditioned. 
Furthermore, the algorithms of \citet{ruan2021linear}, \citet{hanna2023contexts}, and \citet{zhang2025almost} incorporate subroutines tailored to time-varying features.  Even after pruning these components to fit our fixed-feature setting, the implementations remain computationally intensive and become impractical at moderate problem sizes (see Appendix~\ref{sec:additional-experiments}).
Although those routines are justified in broader settings, we find no simple way to streamline them for the plain fixed-feature linear bandit, leaving much of the burden unresolved.

A recent work \citep{ren2024optimal} shows that $\mathcal{O}(\log\log T)$ batches suffice to achieve the regret $\tilde{\mathcal{O}}\!\bigl(\sqrt{dT\log K}\bigr)$ in the \emph{small-$K$ regime} ($\log K \le \mathcal{O}(d)$), the tightest bound known so far.  
However, designing an algorithm that achieves optimal regret in both the \emph{large-$K$} and \emph{small-$K$ regimes} simultaneously remains an open problem. Empirical experiments (\cref{sec:experiment}) further reveal notable weaknesses: the algorithm often eliminates arms prematurely—including the optimal arm—and is highly sensitive to the hyperparameter~$\gamma$, whose choice can enlarge the third batch to $\mathcal{O}\!\bigl((\log T)^{1+\gamma}\bigr)$, exhausting the horizon and impeding exploitation.

In contrast, our proposed algorithm, \texttt{BLAE}, bridges these gaps by attaining minimax-optimal regret guarantees (up to log factors of $T$) in \emph{both} the large-$K$ and small-$K$ regimes, while using only $\Ocal(\log\log T)$ batches. 
We address the common pitfalls of excessive computational overhead and premature arm elimination by introducing a refined regularized G-optimal design approach. 
As a result, \texttt{BLAE} consistently outperforms existing methods in numerical experiments (Section~\ref{sec:experiment}), demonstrating that it is not only theoretically minimax-optimal but also \emph{practically robust}. 

\section{Preliminaries} \label{preliminaries}

\subsection{Notations}
We denote the cardinality of a set by $|\cdot|$, and the indicator function by  $\bbone_A(x)$, where $\bbone_A(x)=1$ if $x \in A$, and $\bbone_A(x)=0$ otherwise. For a vector $x \in \mathbb{R}^d$, the $\ell_2$-norm is denoted by $\|x\|_2$, and the weighted $\ell_2$-norm associated with a positive definite matrix $H$ is defined as $\|x\|_{H} = \sqrt{x^THx}.$ The trace, determinant, and adjugate (the transpose of the cofactor matrix) of $H$ are denoted by $\text{trace}(H)$, $\det(H)$, and $\text{adj}(H)$ respectively. 
We denote by $[n]$ the set $\{1, \ldots, n\}$ for a positive integer $n$, and use $\wedge$ to denote the minimum operator.

\subsection{Problem Setting: Batched Linear Bandits}
\label{sec:batched_linear_bandits}

The stochastic linear bandit setting is a classical framework for sequential decision-making in which an agent aims to maximize cumulative reward by selecting actions (often called \emph{arms}) with unknown but linearly structured rewards. Specifically, the agent maintains a set of $K$ arms $\mathcal{A} := \{x^{(1)}, x^{(2)}, \dots, x^{(K)}\}$,
each embedded in a $d$-dimensional feature space, $x^{(k)} \in \RR^d$ for all $k \in [K]$, and repeatedly chooses which arm to pull over a horizon of $T$ rounds. At each round $t \in [T]$, an arm $x_t \in \mathcal{A}$ is selected, and a noisy reward of the arm $x_t$
\[
r_t = \langle x_t, \theta^* \rangle + \eta_t
\]
is observed, where $\theta^* \in \mathbb{R}^d$ is an unknown parameter vector, and $\eta_t$ is an independent $\sigma$-subgaussian noise with zero mean. The agent’s objective is to minimize the \emph{cumulative expected regret}, defined as
\[
\mathcal{R}(T) = \mathbb{E}\left[\sum_{t=1}^T \bigl(\langle x^*, \theta^*\rangle - \langle x_t, \theta^*\rangle \bigr)\right],
\]
where $x^* \in \argmax_{x \in \mathcal{A}} \,\langle x, \theta^* \rangle$ is an true optimal arm in terms of expected reward.

In fully adaptive linear bandit settings, the policy is typically updated immediately after observing the reward $r_t$ from the selected arm $x_t$, before the next round begins.
However, in many practical applications, such frequent updates can be infeasible due to computational or operational constraints. 
For example, retraining a large-scale recommendation policy after every user interaction may be impractical in terms of both time and resource consumption.
In this work, we focus on the batched variant of linear bandits, where policy updates occur only infrequently, rather than after every round.

\paragraph{Batching as Limited Adaptivity.}
In \textit{batched} linear bandit settings,
the $T$ rounds are structured into $B$ disjoint \emph{batches}: 
\[
\{1, ..., T\} 
= \underbrace{[t_0\!+\!1, \ldots, t_1]}_{\text{Batch 1}} 
\,\cup \dots \,\cup\, 
\underbrace{[t_{B-1}\!+\!1, \ldots, t_B]}_{\text{Batch }B},
\]
where $t_0 = 0$, $t_B = T$, and $B$ is the total number of allowed policy updates. 
A \textit{batch} is thus a contiguous block of rounds during which the agent must use a \emph{fixed} policy.
In other words, within each batch, the agent is not able to update its policy,
effectively deferring feedback from that entire batch until the end. 
Only at the end of batch $\ell$ can the agent update its policy using the rewards collected from rounds $t_{\ell-1}+1$ to $t_\ell$. Consequently, the agent has at most $B - 1$ opportunities to adapt its strategy over the entire horizon, rather than $T - 1$ as in the fully adaptive setting.

\paragraph{Challenges in Batched Decision-Making.}
The restriction to $B$ policy updates (with $B \ll T$) necessitates a more careful balance between exploration and exploitation within each batch. Since the agent cannot adjust its actions mid-batch in response to newly observed rewards, it must design an effective exploration strategy at the beginning of the batch. This strategy involves selecting arms in a way that both collects sufficient information to refine parameter estimates \emph{and} maximizes the immediate rewards.


In the remainder of this paper, we focus on designing a batched linear bandit algorithm with as few updates as possible, while maintaining regret close to the best-known fully adaptive bounds. 

\section{Algorithm: \texttt{BLAE}} \label{algorithm}

We introduce \emph{Batched Linear bandit Algorithm with Elimination} (\texttt{BLAE}), whose pseudocode is provided in Algorithm~\ref{BLAE}. As its name suggests, our algorithm \texttt{BLAE} sequentially eliminates arms and updates the available arm set \(\mathcal{A}_{\ell} = \{x^{(1)},\ldots,x^{(|\mathcal{A}_{\ell}|)}\} \) after completing each batch \(\ell\). 

Before the start of the $\ell$-th batch, the algorithm determines the policy
 $\pi_{\ell}^* := [\pi_{\ell 1}^*, ..., \pi_{\ell |\mathcal{A}_{\ell-1}| }^*]$
over the remaining arms in $\mathcal{A}_{\ell-1}$ 
to construct an effective sampling strategy for that batch (Line 6).
This policy is computed by solving a regularized G-optimal design problem:
\begin{align}\label{eq:G-optimal_design}
V_\ell(\pi_\ell) &\coloneqq  \frac{\lambda}{c_\ell} I + \sum_{i=1}^{|\mathcal{A}_{\ell-1}|}\pi_{\ell i} T^{\frac{2^\ell-1}{2^\ell}} x^{(i)}{x^{(i)}}^T , \notag\\
\pi_\ell^* &\in \underset{\pi_\ell \in \mathcal{D}_\ell}{\text{argmin}} \underset{x \in \mathcal{A}_{\ell-1}}{\max} \|x\|_{V_\ell(\pi_\ell)^{-1}}^2 \; .
\end{align}
Here, $\lambda \in \RR$ and $c_\ell \in (0,1]$ are the regularization parameter and the exploration-exploitation rate for the $\ell$-th batch, respectively. 
The set $\mathcal{D}_\ell$ denotes the space of all probability distributions over the arms in $\mathcal{A}_{\ell-1}$
\[
\mathcal{D_\ell} \coloneqq \bigg\{\pi_\ell \in \mathbb{R}_{\geq 0}^{|\mathcal{A}_{\ell-1}|} \, \bigg| \, \sum_{i=1}^{|\mathcal{A}_{\ell-1}|}\pi_{\ell i}=1\bigg\} \; .
\]
For the $\ell$-th batch, the algorithm is designed to approximately match the batch size $T^{\frac{2^{\ell}-1}{2^{\ell}}}$. 
The eventual probability distribution used for arm allocation is a weighted combination of 
$\pi_{\ell}^*$
and a distribution that assigns full probability to a single arm, given by
\[
\hat{x}_{\ell-1}^* \in \underset{x \in \mathcal{A}_{\ell-2}}{\text{argmax}} \; \langle x, \hat{\theta}_{\ell-1} \rangle \; .
\]
Here, $\hat{x}_{\ell-1}^*$ is the estimated optimal arm based on the observations from the $(\ell-1)$-th batch. If multiple arms satisfy the criterion, one is selected arbitrarily.

Given this probability distribution, the arms are pulled proportionally to their assigned probabilities. For notational convenience, we designate the first arm as the estimated optimal arm from the previous batch among all remaining arms. (Line 8).

Notably, \texttt{BLAE} does not enforce any fixed choice of $c_\ell \in (0,1]$, which provides both theoretical and practical flexibility in balancing exploration and exploitation. In particular, setting $c_1 = 1$ for the first batch is merely a concrete instantiation reflecting the lack of prior information; subsequent batches can adapt $c_\ell$ as more data become available. This design ensures that the presence of $c_\ell$ does not hinder performance or introduce undue complexity.

After pulling arms according to the allocated probabilities, we update the regularized least squares estimator and the estimated optimal arm. (Line 11) Using the estimators  $\hat{\theta}_\ell$  and  $\hat{x}_\ell^*$, we then apply the following elimination rule to remove suboptimal arms (Line 12).
\[
\mathcal{A}_{\ell} = \left\{ x \in \mathcal{A}_{\ell-1} \, \Big| \, \langle \hat{\theta}_{\ell}, \hat{x}_{\ell}^{*} - x \rangle \leq \varepsilon_\ell \right\} \;
\]
Here, while the construction of the elimination threshold $\varepsilon_\ell$ is crucial for our theoretical analysis (its order is specified in \cref{thm:regret of algorithm} and the actual values are specified in Lemma~\ref{lemma:good-event}), to maximize practical performance, we use the bounds without simplification, resulting in complex relational expressions rather than straightforward forms.


Finally, the algorithm reorders the remaining arms in the arm set such that the  estimated optimal arm from the current batch is placed first, ensuring consistency with the indexing convention used in Line 8 (Line 13).

\paragraph{Batched Regularized Least Squares Estimation.} \label{para:batch}
We estimate $\theta^*$ using regularized least squares estimator in a batch-wise manner. After the completion of the $\ell$-th batch, the estimator are updated as follows
\begin{align*}
H_\ell =\!\!\!\sum_{s=t_{\ell-1}+1}^{t_\ell}\!\!\! x_sx_s^T + \lambda I, \quad
\hat{\theta}_{\ell} = H_\ell^{-1}\!\!\!\! \sum_{s=t_{\ell-1}+1}^{t_\ell} x_sr_s \; .
\end{align*}
A naive approach would be to apply the ordinary regularized least squares estimator for all $1\leq s \leq t_\ell$. However, due to policy updates, the noise terms $\eta_s$ for $1\leq s \leq t_{\ell-1}$ influence the arm selection process in the $\ell$-th batch. To derive the concentration bound in Lemma~\ref{lemma:concentration}, we require $H_{\ell}^{-1}$ to be independent of the noise terms $\eta_s$ used for estimator updates. This independence condition is not satisfied in the standard regularized least squares setting, necessitating a refined estimation approach.

While independence is maintained within each batch—i.e., among samples collected before a policy update occurs—we adopt a batched version of regularized least squares estimator to ensure the necessary independence properties. Further mathematical details and the proof can be found in Section~\ref{proof:concentration}.

\begin{algorithm}[tb]
	\caption{\texttt{BLAE}}
        \label{BLAE}
	\begin{algorithmic}[1]
 \STATE {\bfseries Input:} Arm set $\mathcal{A}$, horizon $T$, 
 regularization parameter $\lambda$,
exploration-exploitation rates $\{c_\ell\}_{\ell=1}^{B}$,
 elimination bounds $\{\varepsilon_\ell\}_{\ell=1}^{B}$
 \STATE \hrulefill
 \STATE \textbf{Initialize:} batch $\ell \leftarrow 0$, time $t \leftarrow 0$, arm set $\mathcal{A}_{0} \leftarrow \Acal$
 \WHILE{$t < T$}
 \STATE $\ell \leftarrow \ell + 1$;
 \STATE $\pi_\ell^* \leftarrow {\text{argmin}}_{\pi \in \mathcal{D}_\ell} {\max}_{x \in \mathcal{A}_{\ell-1}} \|x\|_{V_\ell(\pi)^{-1}}^2$ in Eq.\eqref{eq:G-optimal_design};
  \FOR {$i \leftarrow 1,2,\dots,|\mathcal{A}_{\ell-1}|$}
  \STATE Pull arm $x^{(i)}$ repeatedly for the following times\\
   $\Big\{ \lceil T^{\frac{2^\ell-1}{2^{\ell}}}\!\!  \left(c_\ell\pi^*_{\ell i}+(1\!-\!c_\ell)\!\cdot\! \bbone_{\{1\}}\!(i) \right) \rceil \, \wedge\, T\!-\!t \Big\}$;
  \STATE $t\! \leftarrow \!
    \Big\{ \lceil T^{\frac{2^\ell-1}{2^{\ell}}}\!\!  \left(c_\ell\pi^*_{\ell i}+(1\!-\!c_\ell)\!\cdot\! \bbone_{\{1\}}\!(i) \right) \rceil + t\, \wedge\, T \Big\}$;
  \ENDFOR
  \STATE Update batched regularized least squares estimator $\hat{\theta}_\ell$ and estimated optimal arm $\hat{x}^{*}_\ell$;
  \STATE Update $\mathcal{A}_{\ell} \leftarrow \big\{ x \in \mathcal{A}_{\ell-1} \mid \langle \hat{\theta}_{\ell}, \hat{x}_{\ell}^{*} - x \rangle \leq \varepsilon_\ell \big\}$;
  \STATE Rename arms $\mathcal{A}_{\ell} \leftarrow \big\{\hat{x}_{\ell}^*=x^{(1)},x^{(2)},...,x^{(|\mathcal{A}_{\ell}|)}\big\}$;
\ENDWHILE
\end{algorithmic}
\end{algorithm}
\section{Regret Analysis} \label{regret analysis}

As we delve into the regret analysis of our proposed algorithm (\cref{algorithm}), we first present the standard assumptions used in the literature.
\subsection{Assumptions}
\begin{assumption}[Boundedness]
\label{assum:norm}
\(\|x\|_2 \leq 1\) for all $x \in \Acal$ and $\|\theta^*\|_2 \leq 1$.
\end{assumption}
Note that the boundedness is assumed without loss of generality.
If $\|x\|_2 \leq C$ and $\|\theta^*\|_2 \leq C$ for some constant $C$, then the regret bounds would scale by a factor of $C$. This is a standard assumption in the linear bandit literature. 

\begin{assumption}
\label{assum:subgauss}
The noise $\eta_t$ is a 1-subgaussian random variable for all time steps $t \in [T]$.
\end{assumption}
Similar to Assumption~\ref{assum:norm}, the 1-subgaussianity assumption ensures that the regret bounds remain independent of scale for analytical convenience. This is also a standard assumption in the literature.

\subsection{Worst-Case Regret of \texttt{BLAE}}
\begin{theorem}[Regret of \texttt{BLAE}]
\label{thm:regret of algorithm}
Suppose the \texttt{BLAE} algorithm is executed for a total of $T$ rounds with $K$ arms. If the time horizon satisfies $T\geq \Omega(\max\{\frac{K^2}{(d \, \wedge \, \log K)d}, K^{\frac{4}{5}}\})$, and given that $d\geq2$, let
\[\varepsilon_\ell = \Ocal\left(\sqrt{dT^{-\frac{2^\ell-1}{2^\ell}}}\Big(\log(KT) \wedge \sqrt{d+\log T}\Big)\right), \] where $\lambda = \Ocal(1)$ and $c_\ell = \Ocal(1)$, with $c_\ell \in (0,1]$ for $\ell \geq 1$. Then, the worst-case regret of the \texttt{BLAE} algorithm is upper bounded by
\begin{align*}\mathcal{R}(T) &= \Ocal\big( \sqrt{dT}\big(\sqrt{\log(KT)}\wedge \sqrt{d+\log T}\big)\log\log T \big) \\
&= \tilde{\Ocal}\big(\sqrt{dT\log K}\wedge d\sqrt{T}\big)
\end{align*}
\end{theorem}

$\textbf{Discussion of \cref{thm:regret of algorithm}.}$ 
Theorem~\ref{thm:regret of algorithm} shows that \texttt{BLAE} achieves a regret bound of $\widetilde{\Ocal}\bigl(\sqrt{dT\log K}\wedge d\sqrt{T}\bigr)$, matching the best-known (and indeed minimax-optimal) results for fully adaptive linear bandit problems. 
Moreover, \texttt{BLAE} operates with only $\Ocal(\log\log T)$ batches, comparable to the lowest batch complexity among existing batched linear bandit methods. 
Critically, this bound simultaneously covers both the large-$K$ regime ($K \geq \Omega(e^d)$) with $\Ocal(d\sqrt{T})$ regret and the small-$K$ regime ($K \leq \Ocal(e^d)$) with $\Ocal(\sqrt{dT\log K})$ regret, thereby establishing \emph{the tightest known performance guarantees for batched linear bandits} in both regimes. 
To our knowledge, this is the first time such matching lower bounds have been simultaneously achieved under limited adaptivity.

Central to our result are new techniques for regularized G-optimal design under batch updates (see \cref{lemma:regularized-G-optimal-design}), alongside a rigorous concentration bound that handles both batching and regularization (Lemmas~\ref{lemma:concentration} and \ref{lemma:good-event}). 
We also refine the construction of unit-sphere coverings to optimize batch-wise exploration (\cref{lemma 8}). These technical contributions may be of independent interest for other batched (and potentially non-batched) bandit problems.

\subsection{Proof Sketch of Theorem 1}
In this section, we sketch the proof of the worst-case regret bound in Theorem 1 and present key lemmas, with their detailed proofs deferred to the Appendix.

\begin{lemma}[Batch concentration with regularization]
\label{lemma:concentration} After the $\ell$-th batch $(\ell\geq 1)$, the following results holds. For any $x \in \mathbb{R}^d$ and $0<\delta<1$
\begin{equation*}
    \mathbb{P}\left(\langle x, \hat{\theta}_{\ell} - \theta^{*} \rangle \leq \big(\sqrt{2 \log\delta^{-1} } +\sqrt{\lambda}\big)\|x\|_{H_{\ell}^{-1}}\right) \geq 1-\delta\,.
\end{equation*}
\end{lemma}
\cref{lemma:concentration} establishes that the inner product between the estimation error and an arbitrary vector $x$ is upper bounded by the Mahalanobis norm $\|x\|_{H_{\ell}^{-1}}$ with high probability. 
\\
\begin{lemma}[Joint concentration with regularization]
\label{lemma:good-event} Define the following terms 
\begin{align*}
\beta_{\ell1}(\delta) &\coloneqq 2\sqrt{\log\left(\frac{8\pi d(1+\lceil \log_2\log_2T \rceil)^2}{(15/64)^{d-1}\delta^2 }\right)} +2\sqrt{\lambda}\,,\\
\beta_{\ell2}(\delta) &\coloneqq \sqrt{2\log\left(\frac{|\mathcal{A}_{\ell-1}|^2-|\mathcal{A}_{\ell-1}|}{\delta/(1+\lceil \log_2\log_2T \rceil)}\right)} + \sqrt{\lambda} \,,\\
\varepsilon_\ell(\delta) &\coloneqq \underset{x,y \in \mathcal{A}_{\ell-1}}{\max}\|x-y\|_{H_\ell^{-1}} \cdot \big(\beta_{\ell1}(\delta) \wedge \beta_{\ell2}(\delta) \big) \; .
\end{align*}
Here, $\delta$ is a constant within the interval $(0,1)$. Then, the following event holds with probability at least $1-\delta$,
$$\bigcap\limits_{\ell = 1}^{B}\left\{|\langle x-y,\hat{\theta}_\ell - \theta^* \rangle| \leq \varepsilon_\ell(\delta), \; \forall x,y \in \mathcal{A}_{\ell-1}\right\} \; .$$
\end{lemma}
\cref{lemma:good-event} provides a high probability guarantee that the estimation error remains tightly controlled across all batches and arm pairs. The term $\beta_{\ell1}$ quantifies the difference between the batched regularized least squares estimator $\hat{\theta}_{\ell}$ and the unknown parameter $\theta^*$, which plays a crucial role in establishing the regret bound of $\tilde{\Ocal}(d\sqrt{T})$ in the large-$K$ regime. Similarly, $\beta_{\ell2}$ captures the discrepancy between $\hat{\theta}_{\ell}$ and $\theta^*$ from a different perspective, which is essential for deriving the regret bound of $\tilde{\Ocal}(\sqrt{dT\log K})$ in the small-$K$ regime.
By defining $\beta_{\ell}(\delta)\coloneqq \beta_{\ell1}(\delta) \wedge \beta_{\ell2}(\delta)$  as the minimum of these two terms, our framework seamlessly handles both regimes. This motivates the construction of $\varepsilon_\ell(\delta)$ as defined above. For convenience, we define
\[\varepsilon_\ell \coloneqq \underset{x,y \in \mathcal{A}_{\ell-1}}{\max}\|x-y\|_{H_\ell^{-1}} \cdot \beta_\ell\left(T^{-1}\right)\] as it frequently used in regret analysis and also stated in Theorem~\ref{thm:regret of algorithm}. Most existing concentration inequalities focus on individual arms. However, our arm elimination strategy explicitly accounts for the difference between an arm and the estimated optimal arm $\hat{x}_{\ell}^*$. By constructing a concentration inequality based on the difference between arms, we effectively derive a tighter upper bound compared to conventional inequalities that analyze individual arms independently. Intuitively, this approach leverages the fact that $\|x-y\| \leq \|x\|+\|y\|$, enabling a more structured and practical elimination process. Moreover, this result ensures that the optimal arm is preserved throughout the arm elimination process, while suboptimal arms are reliably removed with high probability. Consequently, for all arms available in the $\ell$-th batch, it can be ensured that the suboptimality gap satisfies $\Delta_x = \langle x^*-x,\theta^* \rangle \leq 2\varepsilon_{\ell-1}(\delta)$ with high probability.
\\
\begin{lemma}[Regularized G-optimal design]
\label{lemma:regularized-G-optimal-design}
After the $\ell$-th batch $(\ell\geq 1)$, the following results holds. For an exploration-exploitation rate $c_\ell$ in $(0,1]$,
$$\underset{x\in \mathcal{A}_{\ell-1}}{\max}\|x\|_{H_{\ell}^{-1}}^2 \leq \frac{d}{d\lambda + T^{\frac{2^\ell-1}{2^\ell}}c_\ell} \; .$$
\end{lemma}
\cref{lemma:regularized-G-optimal-design} establishes an upper bound on the uncertainty, measured by the Mahalanobis norm \( \|x\|_{H_\ell^{-1}}^2 \), for any arm in the \(\ell\)-th batch. Most G-optimal design methods used in bandit algorithms omit regularization and rely solely on the Kiefer-Wolfowitz equivalence theorem \citep{kiefer1960equivalence}. However, similar to the argument in Lemma~\ref{lemma:concentration}, under an arm elimination setting, the information matrix $H_\ell$ may not remain invertible without regularization. Thus, a refined analysis that explicitly incorporates the regularization term is necessary. This lemma further demonstrates that we can derive an even tighter upper bound on the uncertainty compared to approaches that do not account for regularization. The exponential decay term \( T^{-\frac{2^\ell - 1}{2^\ell}} \) ensures that uncertainty decreases rapidly as the batch index \(\ell\) increases and the time horizon \( T \) grows, highlighting the algorithm's adaptability to batch-wise updates. By leveraging the bound \( \Delta_x \leq 2\varepsilon_{\ell-1}(\delta) \), which holds for all arms available in the \(\ell\)-th batch with high probability, and applying \cref{lemma:regularized-G-optimal-design}, we can effectively bound the expected cumulative regret as described in \cref{thm:regret of algorithm}.

\section{Why Many Existing Provable Batched Linear Bandit Algorithms Underperform in Practice?} \label{sec:reason of fail}

In this section, we examine why several recently proposed methods—despite strong theoretical guarantees—tend to underperform in empirical evaluations.


A recently proposed and provably efficient algorithm, \emph{Explore-Estimate-Eliminate-Exploit} (\(\texttt{E}^4\)) by \citet{ren2024optimal}, serves as an example of why theoretically sound methods in batched linear bandits underperform in practice. Specifically, \(\texttt{E}^4\) suffers from two critical practical issues that lead to performance deterioration. First, \(\texttt{E}^4\) occasionally eliminates the optimal arm prematurely due to its elimination strategy.
\(\texttt{E}^4\) applies a D-optimal design concentration bound from \citet{lattimore2020bandit}, which assumes independence in the sampling process. However, \(\texttt{E}^4\) does not perform least squares updates in a batch-wise manner, thus it fails to properly account for the independence assumption required for the bound to hold in the batched setting (see Section~\ref{para:batch}). As a result, this elimination strategy---constructed without the necessary constraints---often leads to the premature elimination of the optimal arm.

Second, the  batch sizes of \(\texttt{E}^4\) are highly sensitive to the hyperparameter~$\gamma$, where the third batch size scales with $\Ocal((log T)^{1+\gamma})$. When choosing relatively large values of $\gamma$, as \citet{ren2024optimal} set in their practical set up, the third batch size often exhausts the entire horizon, leaving practically no time to effectively exploit information learned in the third batch. Consequently, the algorithm relies only on information from the first two, relatively short batches. If suboptimal arm identification occurs during these early batches, this leads directly to linear regret.


Other algorithms also encounter significant challenges in practical evaluations of batched linear bandits.  
For instance, \texttt{BatchLinUCB-DG}~\citep{ruan2021linear} splits each batch into two phases: the second phase depends on newly gathered context to refine the policy, a requirement that proves ineffectual when arms are fixed across rounds. Our numerical experiments consistently show that \texttt{BatchLinUCB-DG} struggles to execute meaningful exploration, frequently failing to ``double'' the determinant of its information matrix and thus settling on degenerate (often zero) exploration distributions. In addition, its mixed-softmax policy, which incorporates a G-optimal design at every round, incurs substantial computational overhead, thereby restricting its applicability to larger-scale problems.

Algorithm~3 in \citet{hanna2023contexts} addresses yet another dimension of complexity by transforming the contextual bandit into a non-contextual one through constructing a \(\frac{1}{T}\)-net \(\Theta'\) over a bounded parameter space. Although the authors claim \(|\Theta'|\le (6T)^d,\) the algorithm must calculate 
\(\arg\max_{a\in \mathcal{A}_t} \langle a,\theta\rangle\)
for each \(\theta\in\Theta'\) at each batch step, creating tremendous computational burdens that far exceed what is tractable for real-world applications.

Lastly, Algorithm~1 in \citet{zhang2025almost} adopts a two-phase batch-splitting strategy similar to that of \citet{ruan2021linear}, thereby inheriting the similar limitations. As with its counterpart, the algorithm produces suboptimal policies in fixed-arm environments and incurs considerable computational overhead from computing a G-optimal design at every round of the first batch, a cost that grows rapidly when the action set is large or the time horizon $T$ increases.

Our runtime experiments in Table~\ref{table:runtime-with-Ruan-and-Hanna} confirm that these approaches~\citep{ruan2021linear, hanna2023contexts, zhang2025almost} struggle to scale efficiently even for moderately sized $T$ and other problem parameters, 
rendering them impractical for large-scale problem instances. Moreover, all three methods~\citep{ruan2021linear, hanna2023contexts, zhang2025almost} assume that contexts are drawn independently and identically distributed (i.i.d.). In fixed-arm environments, the contexts remain identically distributed but are not independent, violating this assumption and undermining the theoretical validity of these methods in our setting. 

More broadly, the combination of premature arm elimination, extensive computational costs, and the lack of consideration for independence requirements in batched settings helps explain why methods that appear theoretically sound may struggle to deliver in real-world applications.

\section{Numerical Experiments}\label{sec:experiment}

\begin{figure*}[ht]
\centering
\begin{subfigure}{\textwidth}
\includegraphics[width=\textwidth]{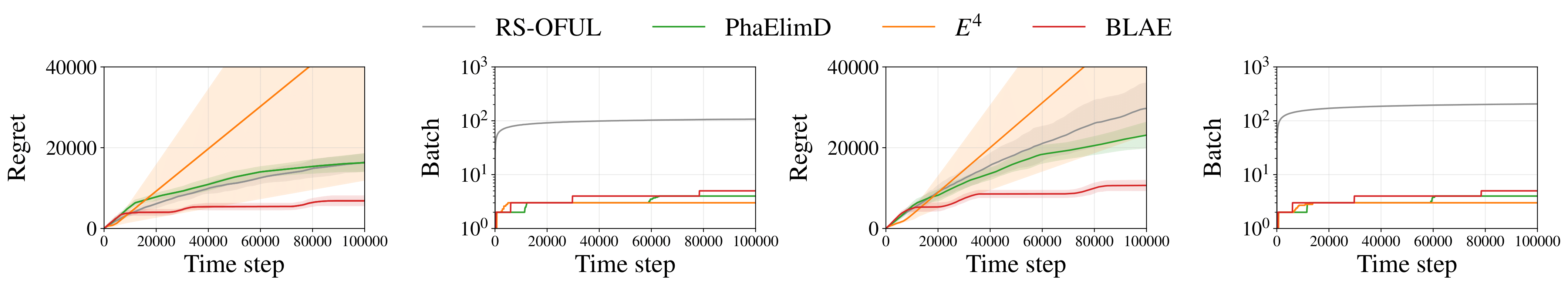}
\vspace{-0.7cm}
    \begin{center}
        \makebox[0.25\textwidth][c]{(a) $K = 50, d=5$} 
        \makebox[0.25\textwidth][c]{}
        \makebox[0.25\textwidth][c]{(b) $K = 50, d=10$}
    \end{center}
\end{subfigure}
\vspace{0.3cm}
\begin{subfigure}{\textwidth}
\includegraphics[width=\textwidth]{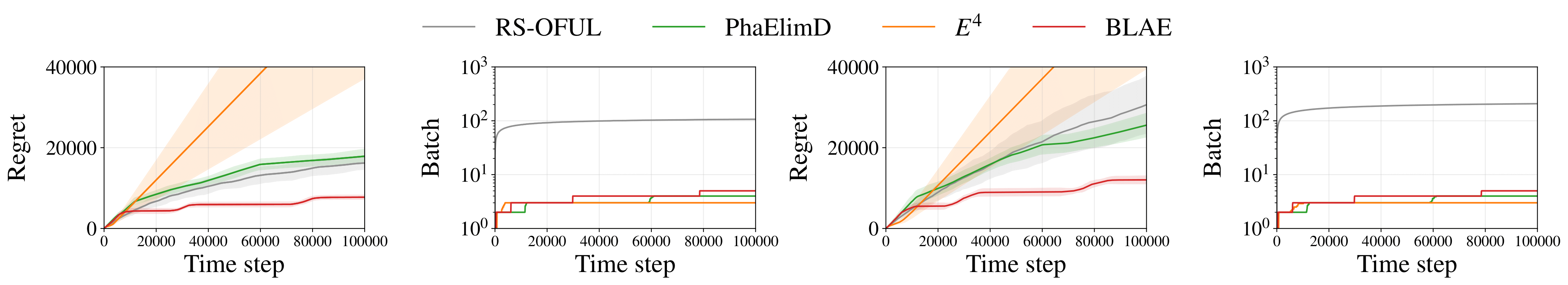}
\vspace{-0.7cm}
    \begin{center}
        \makebox[0.25\textwidth][c]{(c) $K = 100, d=5$} 
        \makebox[0.25\textwidth][c]{}
        \makebox[0.25\textwidth][c]{(d) $K = 100, d=10$}
    \end{center}
\end{subfigure}
\begin{subfigure}{\textwidth}
\includegraphics[width=\textwidth]{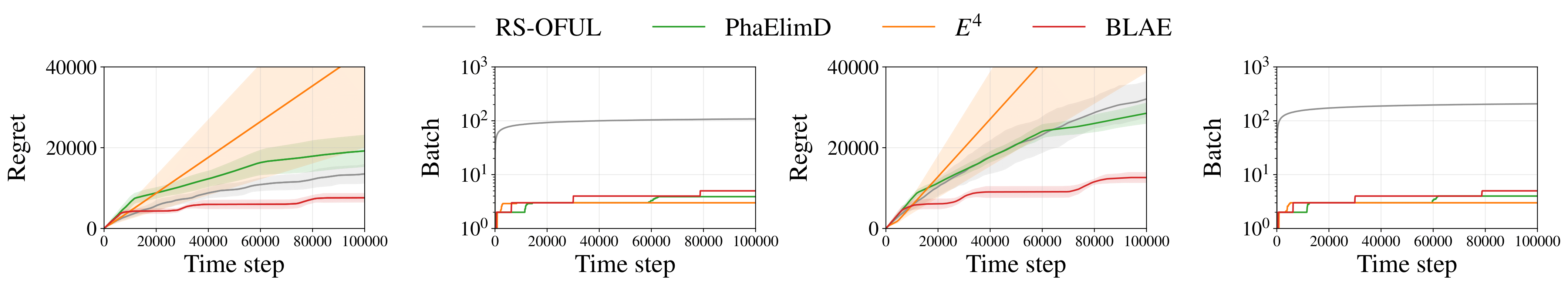}
\vspace{-0.7cm}
    \begin{center}
        \makebox[0.25\textwidth][c]{(e) $K = 200, d=5$} 
        \makebox[0.25\textwidth][c]{}
        \makebox[0.25\textwidth][c]{(f) $K = 200, d=10$}
    \end{center}
\end{subfigure}
\begin{subfigure}{\textwidth}
\includegraphics[width=\textwidth]{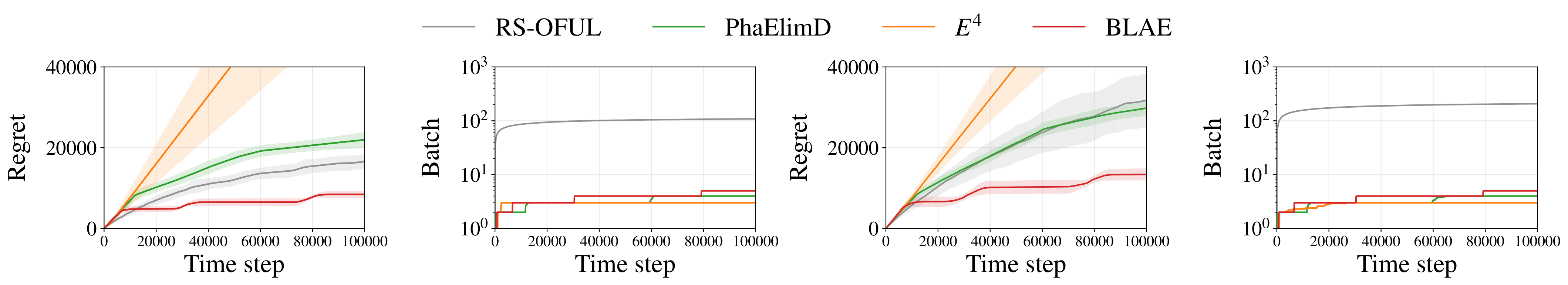}
\vspace{-0.7cm}
    \begin{center}
        \makebox[0.25\textwidth][c]{(g) $K = 400, d=5$} 
        \makebox[0.25\textwidth][c]{}
        \makebox[0.25\textwidth][c]{(h) $K = 400, d=10$}
    \end{center}
\end{subfigure}
\setcounter{figure}{0}
\captionof{figure}{Regret and batch complexity over time for different values of $K$ and $d$.}
\label{uniform figure}
\end{figure*}

\subsection{Experimental Setup}
In this section, we empirically evaluate the performance of the \texttt{BLAE} algorithm. We measure the cumulative regret over $T=100,000$ time steps. The $K$ arms and \(\theta^*\) are randomly sampled from a \(d\)-dimensional uniform distribution. The experiment is repeated 10 times, with the arms being independently resampled from the uniform distribution in each run. We conduct experiments with the following $(K,d)$ parameter pairs $(K,d) = (50,5), (50,10),$ $(100,5),(100,10),(200,5),(200,10), (400,5), (400,10)$.

We compare the performance of \texttt{BLAE} with the previously known state-of-the-art algorithms: \textit{Explore-Estimate-Eliminate-Exploit} (denoted by $\texttt{E}^4$; \citealt{ren2024optimal}), \textit{Rarely Switching OFUL} (denoted by \texttt{RS-OFUL}; \citealt{abbasi2011improved}), and the \textit{phased elimination algorithm with D-optimal design} (denoted by \texttt{PhaElimD}; \citealt{esfandiari2021regret,lattimore2020bandit}). For \texttt{BLAE}, we set the parameters as $\lambda = 1$ and $c_\ell = |\mathcal{A}_{\ell-1}|/|\mathcal{A}_0|$ for $\ell \geq 1$. This choice allows $c_\ell$ to adaptively reflect the level of uncertainty based on the number of remaining arms, ensuring both flexibility and practical applicability. For $\texttt{E}^4$, we configure the parameters identically to the practical setup in \citet{ren2024optimal}. For \texttt{RS-OFUL}, we set the switching parameter to $C=0.5$. For \texttt{PhaElimD}, we design the batch sizes as $T^{1-2^{-i}}$ instead of $q^i$ to achieve better performance.

Algorithms proposed by \citet{ruan2021linear,hanna2023contexts,zhang2025almost} incur significant computational overhead; their runtimes are reported in Section~\ref{sec:additional-experiments}, as explanation for their empirical inefficiency is presented in~\cref{sec:reason of fail}. Accordingly, we omit regret plots for these methods.

\subsection{Experimental Results}

For the experimental results, we present two types of figures to summarize our evaluations. First, the average of cumulative regret (solid line) with its standard deviation (shaded region) is plotted based on 10 independent runs. 
Second, the average batch complexity (solid line) is depicted using the same 10 trials, illustrating how frequently each algorithm updates its policy.

As illustrated in Figure~\ref{uniform figure}, \texttt{BLAE} consistently and significantly outperforms all other algorithms, exhibiting both low cumulative regret and notable stability—surpassing even \texttt{RS-OFUL}, which executes over 100 parameter updates, far more than \texttt{BLAE}. 
These outcomes confirm that \texttt{BLAE} not only achieves the tightest known theoretical guarantees but also delivers strong empirical performance, thereby fulfilling its design objectives of near-optimal regret and minimal batch complexity.


Among recently proposed methods, 
$\texttt{E}^4$~\citep{ren2024optimal} is shown to attain minimal batch complexity (Figure~\ref{uniform figure}). 
This advantage, however, comes at a cost in regret performance: the algorithm records the highest cumulative regret and shows marked instability.  Minimizing the number of batches is therefore of limited value if the resulting regret is unsustainable.  
Across 80 independent experiments, \(\texttt{E}^4\) shows substantial performance variability due to its aggressive arm‐elimination policy and sensitivity to batch size. As noted in \cref{sec:reason of fail}, its elimination strategy fails to respect independence in the batched setting, occasionally discarding the optimal arm prematurely. In addition, the algorithm’s sensitivity to the hyperparameter~$\gamma$ often causes the third batch to consume almost the entire horizon, leaving little opportunity for exploitation. Our evaluation used the official implementation released by \citet{ren2024optimal}.


Overall, the superior empirical results of \texttt{BLAE} validate its unique advantage: it couples a proven minimax-optimal regret bound with practical efficiency. This combination of theoretical optimality and consistently strong performance sets \texttt{BLAE} apart from prior batched linear bandit methods. 

\section{Conclusion}
We present a practical and minimax-optimal algorithm for batched linear bandits that unifies theoretical tightness and empirical efficiency. By integrating arm elimination with a regularized G-optimal design, our method achieves $\widetilde{\Ocal}\bigl(\sqrt{dT\log K} \,\wedge\, d\sqrt{T}\bigr)$ regret in only $\Ocal(\log\log T)$ batches, matching known lower bounds for both large-$K$ and small-$K$ regimes. 
Extensive numerical evaluations confirm its superiority over existing methods, illustrating how 
our method bridges the gap between provable efficiency and empirical performance. 

\section*{Impact Statement}
This paper presents work whose goal is to advance the field of Machine Learning. There are many potential societal consequences of our work, none which we feel must be specifically highlighted here.

\section*{Acknowledgements}
This work was supported by the National Research Foundation of Korea~(NRF) grant funded by the Korea government~(MSIT) (No.  RS-2022-NR071853 and RS-2023-00222663), by Institute of Information \& communications Technology Planning \& Evaluation~(IITP) grant funded by the Korea government~(MSIT) (No. RS-2025-02263754), and by AI-Bio Research Grant through Seoul National University.

\bibliography{references.bib}
\bibliographystyle{icml2025}

\newpage
\appendix
\onecolumn
\renewcommand\thesection{\Alph{section}}
\setcounter{section}{0}
\textbf{\LARGE Appendix}
\section{Proof of Lemma 1}
The proof of \cref{lemma:concentration} relies on the following two technical lemmas. Here, we assume that subgaussian random variables have zero mean.
\begin{lemma}
\label{lemma 4} Let $X$ be a $\sigma$-subgaussian random variable. Then for any $\varepsilon \geq 0$, the following concentration bound holds 
\[\mathbb{P}(X \geq \varepsilon) \leq e^{-\frac{\varepsilon^2}{2\sigma^2}}.\]
\end{lemma}
\begin{proof} Let $\lambda > 0$. By Markov's inequality and the definition of subgaussianity, we have 
\[\mathbb{P}(X\geq \varepsilon) = \mathbb{P}(e^{\lambda X} \geq e^{\lambda \varepsilon})\leq \mathbb{E}[e^{\lambda X}]e^{-\lambda\varepsilon}\leq e^{\frac{\lambda^2\sigma^2}{2}-\lambda\varepsilon}. \]
To optimize the upper bound, we set $\lambda = \frac{\varepsilon}{\sigma^2}$. Substituting this value of $\lambda$, we obtain the desired inequality. 
\end{proof}
\begin{lemma}
\label{lemma 5} Let $X$ be a $\sigma$-subgaussian random variable, and let $X_1$ and $X_2$ be independent $\sigma_1$-subgaussian and $\sigma_2$-subgaussian random variables, respectively. Then, the following properties hold
\begin{enumerate}[label=(\alph*)]
\item For any $c \in \mathbb{R}$, $cX$ is $|c|\sigma\text{-subgaussian}$

\item $X_1 + X_2$ is $\sqrt{\sigma_{1}^2+\sigma_{2}^2}\text{-subgaussian.}$
\end{enumerate}
\end{lemma} 
\begin{proof} By the definition of subgaussianity, for any $\lambda \in \mathbb{R}$, we have $\mathbb{E}[e^{\lambda X}] \leq e^{\frac{\lambda^2 \sigma^2}{2}}$. Substituting $\lambda$ with $c\lambda$ proves (a) as follows
\[\mathbb{E}[e^{\lambda (cX)}] = \mathbb{E}[e^{(c\lambda) X}]\leq e^{\frac{(c\lambda)^2 \sigma^2}{2}} = e^{\frac{\lambda^2 (|c|\sigma)^2}{2}} .\]

By the definition of subgaussianity, for any $\lambda \in \mathbb{R}$, we have $\mathbb{E}[e^{\lambda X_1}] \leq e^{\frac{\lambda^2 \sigma_1^2}{2}}$ and $\mathbb{E}[e^{\lambda X_2}] \leq e^{\frac{\lambda^2 \sigma_2^2}{2}}$. Using the independence of $X_1$ and $X_2$, we can write
\[\mathbb{E}[e^{\lambda (X_1+X_2)}] = \mathbb{E}[e^{\lambda X_1}]\mathbb{E}[e^{\lambda X_2}] \leq e^{\frac{\lambda^2 \sigma_1^2}{2}}e^{\frac{\lambda^2 \sigma_2^2}{2}} = e^{\frac{\lambda^2(\sigma_1^2+\sigma_2^2)}{2}} .\]
Thus, (b) is also proved.
\end{proof}

\begin{proof}[Proof of Lemma 1] \label{proof:concentration} For any $x \in \mathbb{R}^d$, substituting $\hat{\theta}_\ell$ with the batched regularized least squares
estimator and expanding \(r_s\), we obtain
\begin{align*}
\langle x, \hat{\theta}_{\ell} - \theta^{*} \rangle &= \langle x,H_{\ell}^{-1}\sum_{s=t_{\ell-1}+1}^{t_\ell} x_sr_s - \theta^{*} \rangle \\
&= \langle x,H_{\ell}^{-1} \sum_{s=t_{\ell-1}+1}^{t_\ell} x_s(x_{s}^{T}\theta^{*}+\eta_{s}) - \theta^{*} \rangle \\
&= \langle x,H_{\ell}^{-1} (\sum_{s=t_{\ell-1}+1}^{t_\ell} x_s\eta_{s} + (H_{\ell} - \lambda I)\theta^*) - \theta^{*} \rangle \\
&= \langle x,H_{\ell}^{-1} \sum_{s=t_{\ell-1}+1}^{t_\ell} x_s\eta_{s} - \lambda H_{\ell}^{-1}\theta^* \rangle \\
&= \sum_{s=t_{\ell-1}+1}^{t_\ell} \langle x,H_{\ell}^{-1}x_s \rangle \eta_{s} - \lambda \langle x,H_{\ell}^{-1}\theta^* \rangle \; . \tag{1}
\end{align*}
\\
The matrix $H_{\ell}$ is independent of the noise terms $\{\eta_{s}\}_{s=t_{\ell-1}+1}^{t_\ell}$. Furthermore, these $1$-subgaussian random variables $\{\eta_{s}\}_{s=t_{\ell-1}+1}^{t_\ell}$ are mutually independent. According to \cref{lemma 5}, the term
$$\sum_{s=t_{\ell-1}+1}^{t_\ell} \langle x, H_{\ell}^{-1}x_s \rangle \eta_{s} \quad \text{is} \quad\sqrt{\sum_{s=t_{\ell-1}+1}^{t_\ell} \langle x, H_{\ell}^{-1}x_s \rangle^2} -\text{subgaussian}.$$
\\
Applying the concentration inequality from \cref{lemma 4} to this subgaussian random variable yields
\begin{equation}
\mathbb{P}\left(\sum_{s=t_{\ell-1}+1}^{t_\ell} \langle x,H_{\ell}^{-1}x_s \rangle \eta_{s} \leq \sqrt{2\log\left(\frac{1}{\delta}\right)\sum_{s=t_{\ell-1}+1}^{t_\ell} \langle x,H_{\ell}^{-1}x_s \rangle^{2}}\,\right) \geq 1-\delta \; . \tag{2}
\end{equation}
\\
For the second term in (1), we apply the Cauchy-Schwarz inequality, yielding
\begin{equation}
|\langle x,H_{\ell}^{-1}\theta^* \rangle| \leq \|x\|_{H_{\ell}^{-2}} \|\theta^*\| \leq \|x\|_{H_{\ell}^{-2}} \leq \frac{\|x\|_{H_{\ell}^{-1}}}{\sqrt{\lambda}} \; . \tag{3} 
\end{equation}
Combining inequalities (2) and (3) with equation (1), we obtain
$$\mathbb{P}\left(\langle x, \hat{\theta}_{\ell}- \theta^{*} \rangle \leq \sqrt{2\log\left(\frac{1}{\delta}\right)\sum_{s=t_{\ell-1}+1}^{t_\ell} \langle x,H_{\ell}^{-1}x_s \rangle^{2}} + \sqrt{\lambda} \|x\|_{H_{\ell}^{-1}}\right) \geq 1-\delta \; .$$
To obtain a more concise bound, we observe that
$$\sum_{s=t_{\ell-1}+1}^{t_\ell} \langle x,H_{\ell}^{-1}x_s \rangle^{2} \leq x^{T}H_{\ell}^{-1}\left(\sum_{s=t_{\ell-1}+1}^{t_\ell} x_{s}x_{s}^{T} + \lambda I \right)H_{\ell}^{-1}x = x^{T}H_{\ell}^{-1}H_{\ell}H_{\ell}^{-1}x=\|x\|_{H_{\ell}^{-1}}^2 \; .$$
This completes the proof of \cref{lemma:concentration}:
$$\mathbb{P}\left(\langle x, \hat{\theta}_{\ell} - \theta^{*} \rangle \leq \bigg(\sqrt{2\log\left(\frac{1}{\delta}\right)} +\sqrt{\lambda}\bigg)\|x\|_{H_{\ell}^{-1}}  \right) \geq 1-\delta \; . \;$$
\end{proof}
\section{Proof of Lemma 2}
The proof of \cref{lemma:good-event} relies on the following five technical lemmas.
\begin{lemma}
\label{lemma 6}\cite{li2010concise} Let $A_d$ denote the surface area of an \(d\)-dimensional hypersphere of radius 1, and let $A_d^{\text{cap}}$ denote the surface area of a hyperspherical cap on an \(d\)-dimensional sphere of radius 1. Then, the following result holds for \( \phi \in (0, \frac{\pi}{2}] \), where \( I_x(a, b) \) is the regularized incomplete beta function
\[
A_d^{\text{cap}} = \frac{A_d}{2} I_{\sin^2\phi}\left(\frac{d-1}{2}, \frac{1}{2}\right) \; .
\]
\end{lemma}
\begin{lemma}
\label{lemma 7}\cite{abeille2017linear} For any integer \(d \geq 2\),
\[
\frac{\Gamma (\frac{d+1}{2})}{\Gamma (\frac{d}{2}+1)} \leq \sqrt{\frac{2}{d}} \; .
\]
\end{lemma}
\begin{lemma}
\label{lemma 8} For any $\zeta \in (0,1)$, there exists a finite cover $C_\zeta \subset S^{d-1}$ such that for all $x \in S^{d-1}$, there exists $y \in C_\zeta$ with $\|x - y\| \leq \zeta$, and
$|C_\zeta| \leq \sqrt{2\pi d} \cdot (\zeta^2 - \zeta^4/4)^{-\frac{d-1}{2}}$. Here, $S^{d-1} \coloneqq \{x \in \mathbb{R}^d : \|x\| = 1\}$ denotes the $(d-1)$-dimensional unit sphere in $\mathbb{R}^d$, and $d \geq 2$.
\end{lemma}
\begin{proof} Let $\mathcal{X} \subset \mathbb{R}^d.$ A subset $C \subset \mathcal{X}$ is called an $\zeta$-cover of $\mathcal{X}$ if $\mathcal{X} \subset \bigcup_{x \in C}B_{\zeta}(x).$ Similarly, an $\zeta$-packing of $\mathcal{X}$ is a subset $P \subset \mathcal{X}$ such that for any distinct $x,y \in P,$ $\|x-y\|>\zeta.$
We define $N(\zeta)$ as the minimum cardinality of an $\zeta$-cover of $S^{d-1}$, and $M(\zeta)$ as the maximum cardinality of an $\zeta$-packing of $S^{d-1}$. Specifically,
\[
N(\zeta) \coloneqq \min\{|C| : C \text{ is an \(\zeta\)-cover of } S^{d-1} \}, \quad M(\zeta) \coloneqq \max\{|P| : P \text{ is an \(\zeta\)-packing of } S^{d-1} \} \; .
\]
Now, let $P$ be an arbitrary $\zeta$-packing of $S^{d-1}.$ For any distinct $x,y \in P$, we have $B_{\frac{\zeta}{2}}(x) \cap B_{\frac{\zeta}{2}}(y) = \emptyset$. \\
Therefore, we can write
\[\underset{x \in P}{\bigsqcup}B_{\frac{\zeta}{2}}(x) \cap S^{d-1} \subset S^{d-1} \; . \tag{4}\]
\\
From (4) and \cref{lemma 6}, and the arbitrariness of $\zeta$-packing $P$, we obtain
$$M(\zeta) \leq \frac{A_d}{A_d^{\text{cap}}} = \frac{2}{I_{\sin^2(2\sin^{-1}\frac{\zeta}{2})}\left(\frac{d-1}{2}, \frac{1}{2}\right)} = \frac{2}{I_{\zeta^2(1-\frac{\zeta^2}{4})}\left(\frac{d-1}{2}, \frac{1}{2}\right)} \; .$$
\\
Using the monotonicity of \(t \rightarrow (1-t)^{-\frac{1}{2}}\) on \((0,\frac{3}{4})\), along with \(0 < \zeta^2(1-\frac{\zeta^2}{4}) < \frac{3}{4}\) for \(\zeta \in (0,1)\), and applying \cref{lemma 7}, we derive
\begin{align*}
\frac{2}{I_{\zeta^2(1-\frac{\zeta^2}{4})}\left(\frac{d-1}{2}, \frac{1}{2}\right)} &= \frac{2B(\frac{d-1}{2},\frac{1}{2})}{B_{\zeta^2(1-\frac{\zeta^2}{4})}(\frac{d-1}{2},\frac{1}{2})} \\
&= \frac{2B(\frac{d-1}{2},\frac{1}{2})}{\int_0^{\zeta^2(1-\frac{\zeta^2}{4})} t^{\frac{d-3}{2}}(1-t)^{-\frac{1}{2}}dt} \\
&\leq \frac{2B(\frac{d-1}{2},\frac{1}{2})}{\int_0^{\zeta^2(1-\frac{\zeta^2}{4})} t^{\frac{d-3}{2}}dt} \\
&= \frac{(d-1)B(\frac{d-1}{2},\frac{1}{2})}{(\zeta^2-\frac{\zeta^4}{4})^{\frac{d-1}{2}}} \\
&= \frac{(d-1)\sqrt{\pi}}{(\zeta^2-\frac{\zeta^4}{4})^{\frac{d-1}{2}}} \cdot \frac{\Gamma(\frac{d-1}{2})}{\Gamma(\frac{d}{2})} \\
&= \frac{d\sqrt{\pi}}{(\zeta^2-\frac{\zeta^4}{4})^{\frac{d-1}{2}}} \cdot \frac{\Gamma(\frac{d+1}{2})}{\Gamma(\frac{d}{2}+1)} \\
&\leq \frac{\sqrt{2\pi d}}{(\zeta^2-\frac{\zeta^4}{4})^{\frac{d-1}{2}}} \; .
\end{align*}
\\
Here, \(B_x(a,b)\) is the incomplete beta function and \(B(a,b)\) is the beta function.
\\
Since $M(\zeta) \leq \sqrt{2\pi d} \,(\zeta^2-\frac{\zeta^4}{4})^{-\frac{d-1}{2}} < \infty,$ there exists a maximal $\zeta$-packing $P_\zeta$ of $S^{d-1}.$ In this case, $P_\zeta$ is also an $\zeta$-cover of $S^{d-1}$, implying $N(\zeta)\leq M(\zeta).$
\\
Thus, combining these results completes the proof:
\[
N(\zeta) \leq M(\zeta) \leq  \frac{\sqrt{2\pi d}}{(\zeta^2-\frac{\zeta^4}{4})^{\frac{d-1}{2}}}\; .
\]
\end{proof}

\begin{lemma}
\label{lemma 9} After the $\ell$-th batch $(\ell \geq 1)$, the following bound holds for any $0 < \zeta < 1$ and $d \geq 2$ with probability at least $1-\delta$
\[
\|\hat{\theta}_{{\ell}}-\theta^{*} \|_{V_{{\ell}}} \leq \frac{\sqrt{(d-1)\log(\frac{4}{4\zeta^2-\zeta^4})+2\log(\frac{\sqrt{2\pi d}}{\delta})} +\sqrt{\lambda}}{1-\zeta}\; .
\]
\end{lemma}
\begin{proof} If $\hat{\theta}_\ell = \theta_*$, the lemma is trivial. Thus, we only consider the case where $\hat{\theta}_\ell \neq \theta_*$. 
\\
Let $X = \frac{H_{\ell}^{\frac{1}{2}}(\hat{\theta}_{{\ell}} - \theta_{*})}{\|\hat{\theta}_{{\ell}} - \theta_{*}\|_{H_{\ell}}}.$ Since $\|X\|=1$, it follows that $X \in S^{d-1}$, where $S^{d-1}$ is the unit sphere in $\mathbb{R}^d$.
\\
Let $C_{\zeta} \subset S^{d-1}$ be the \(\zeta\)-covering set given in \cref{lemma 8}, and define the event
$$E \coloneqq \left\{ \exists \, x \in C_\zeta \; s.t. \; \langle H_{\ell}^{\frac{1}{2}}x, \hat{\theta}_{\ell} - \theta_{*} \rangle > \sqrt{2\log\left(\frac{|C_\zeta|}{\delta}\right)} + \sqrt{\lambda} \right\} \; .$$
Using the property $\|H_{\ell}^{\frac{1}{2}}x\|_{H_{\ell}^{-1}} = \|x\| = 1$ for all $x \in S^{d-1}$, and by applying \cref{lemma:concentration}, we have $\mathbb{P}(E) \leq \delta$.
\\
When the event $E$ does not occur, the following inequality holds
\begin{align*}
\|\hat{\theta}_{\ell} - \theta_{*}\|_{H_{\ell}} &= \max_{x \in S^{d-1}} \langle H_{\ell}^{\frac{1}{2}}x, \hat{\theta}_{\ell} - \theta_{*} \rangle \\
&= \max_{x \in S^{d-1}}\min_{y \in C_\zeta} \langle H_{\ell}^{\frac{1}{2}}(x-y), \hat{\theta}_{\ell} - \theta_{*} \rangle + \langle H_{\ell}^{\frac{1}{2}}y, \hat{\theta}_{\ell} - \theta_{*} \rangle \\
&\leq \max_{x \in S^{d-1}}\min_{y \in C_\zeta} \|\hat{\theta}_{\ell} - \theta_{*}\|_{H_{\ell}} \|x-y\| + \langle H_{\ell}^{\frac{1}{2}}y, \hat{\theta}_{\ell} - \theta_{*} \rangle \\
&\leq \zeta \|\hat{\theta}_{\ell} - \theta_{*}\|_{H_{\ell}} + \sqrt{2\log\left(\frac{|C_\zeta|}{\delta}\right)} +\sqrt{\lambda} \\
&\leq \zeta \|\hat{\theta}_{\ell} - \theta_{*}\|_{H_{\ell}} + \sqrt{(d-1)\log\left(\frac{4}{4\zeta^2-\zeta^4}\right)+2\log\left(\frac{\sqrt{2\pi d}}{\delta}\right)} +\sqrt{\lambda} \; .
\end{align*}
Therefore, we prove \cref{lemma 9}:
\[
\mathbb{P}\left(\|\hat{\theta}_{{\ell}}-\theta^{*} \|_{H_{{\ell}}} \leq \frac{\sqrt{(d-1)\log(\frac{4}{4\zeta^2-\zeta^4})+2\log(\frac{\sqrt{2\pi d}}{\delta})} +\sqrt{\lambda}}{1-\zeta}\, \right) \geq 1-\delta \; .
\]
\end{proof}
Here, $\zeta \in (0,1)$ is a tunable parameter. Although one could optimize over $\zeta$, we fix $\zeta = 0.5$ for simplicity, as this choice affects the results only up to a constant factor. We nevertheless retain $\zeta$ as a symbolic variable in the subsequent analysis to preserve generality; readers may safely substitute $\zeta = 0.5$ throughout. For convenience, we define 
\[
M_\ell \coloneqq \, \underset{x,y \in \mathcal{A}_{\ell-1}}{\max}\|x-y\|_{H_\ell^{-1}}^2 \; .
\]
\begin{lemma}
\label{lemma 10} After the $\ell$-th batch $(\ell\geq 1)$, the following result holds for all $x,y \in \mathcal{A}_{\ell-1}$ with probability at least $1-\delta$
$$|\langle x-y, \hat{\theta}_{\ell}-\theta^{*} \rangle| \leq \sqrt{M_\ell} \cdot \beta_\ell(\delta+\delta\lceil \log_2\log_2T \rceil) \; .$$
\end{lemma}
\begin{proof} Using the Cauchy-Schwarz inequality and \cref{lemma 9}, the following result holds for all $x,y \in \mathcal{A}_{\ell-1}$ with probability at least $1-\frac{\delta}{2}$
\[
|\langle x-y, \hat{\theta}_{\ell} - \theta^* \rangle| \leq \|x-y\|_{H_{{\ell}}^{-1}}\|\hat{\theta}_{\ell} - \theta^*\|_{H_{{\ell}}} \leq \frac{\sqrt{\|x-y\|_{H_{{\ell}}^{-1}}^2 }\left(\sqrt{(d-1)\log(\frac{4}{4\zeta^2-\zeta^4})+2\log(\frac{2\sqrt{2\pi d}}{\delta})} +\sqrt{\lambda}\right)}{1-\zeta} \; . 
\]
Since $\|x-y\|_{H_{\ell}^{-1}}^2 \leq M_{\ell}$, we obtain
\[\mathbb{P}\left(
|\langle x-y, \hat{\theta}_{\ell} - \theta^* \rangle| \leq \frac{\sqrt{M_{\ell}}\left(\sqrt{(d-1)\log(\frac{4}{4\zeta^2-\zeta^4})+2\log(\frac{2\sqrt{2\pi d}}{\delta})} +\sqrt{\lambda}\right)}{1-\zeta}, \; \forall x,y \in \mathcal{A}_{\ell-1} \right) \geq 1-\frac{\delta}{2} \; . \tag{5}
\]
Next, by applying \cref{lemma:concentration} to the \(\binom{|\mathcal{A}_{\ell-1}|}{2}\) pairs of $x,y \in \mathcal{A}_{\ell-1}$ such that $\langle x-y, \hat{\theta}_{\ell} - \theta^* \rangle \geq 0$, we know that for all \(x,y \in \mathcal{A}_{\ell-1}\), the following holds with probability at least \(1-\frac{\delta}{2}\)
$$|\langle x-y, \hat{\theta}_{\ell}-\theta^{*} \rangle| \leq  \left(\sqrt{2\log\left(\frac{|\mathcal{A}_{\ell-1}|^2-|\mathcal{A}_{\ell-1}|}{\delta}\right)}+ \sqrt{\lambda}\right) \|x-y\|_{H_{\ell}^{-1}} \; .$$
Since $\|x-y\|_{H_{\ell}^{-1}}^2 \leq M_{\ell}$, we obtain
\[\mathbb{P}\left(|\langle x-y, \hat{\theta}_{\ell}-\theta^{*} \rangle| \leq  \sqrt{M_\ell}\left(\sqrt{2\log\left(\frac{|\mathcal{A}_{\ell-1}|^2-|\mathcal{A}_{\ell-1}|}{\delta}\right)}+ \sqrt{\lambda} \right), \; \forall x,y \in \mathcal{A}_{\ell-1} \right) \geq 1-\frac{\delta}{2} \; . \tag{6}
\]
Using the definition of \(\beta_\ell(\delta)\), we calculate $\beta_{\ell}(\delta+\delta\lceil \log_2\log_2 T \rceil)$ as
\[
\beta_{\ell}(\delta+\delta\lceil \log_2\log_2 T \rceil) = \left(\frac{\sqrt{(d-1)\log(\frac{4}{4\zeta^2-\zeta^4})+2\log(\frac{2\sqrt{2\pi d}}{\delta})} +\sqrt{\lambda}}{1-\zeta} \, \bigwedge \, \sqrt{2\log\left(\frac{|\mathcal{A}_{\ell-1}|^2-|\mathcal{A}_{\ell-1}|}{\delta}\right)} + \sqrt{\lambda}\right) \; .
\]
Combining the result from (5) and (6), we prove \cref{lemma 10}:
\[
\mathbb{P}\left(|\langle x-y, \hat{\theta}_{\ell}-\theta^{*} \rangle| \leq \sqrt{M_\ell} \cdot \beta_\ell(\delta+\delta\lceil \log_2\log_2T \rceil), \; \forall x,y \in \mathcal{A}_{\ell-1}\right) \geq 1-\delta \; .
\]
\end{proof}
\begin{proof}[Proof of Lemma 2] At the $\lceil \log_2\log_2 T \rceil$-th and $(1+\lceil \log_2\log_2 T \rceil)$-th batch, the number of arm pulls satisfies 
\[
T^{1-\frac{1}{2^{1+\lceil \log_2\log_2 T \rceil}}} \geq T^{1-\frac{1}{2^{\lceil \log_2\log_2 T \rceil}}} \geq T^{1-\frac{1}{\log_2T}} = \frac{T}{2} \; .
\]
Thus, the total number of batches $B$ is bounded by
$$B \leq 1+\lceil \log_2\log_2 T \rceil \; .$$
Let $E_{\ell}\coloneqq \{|\langle x-y,\hat{\theta}_\ell - \theta^* \rangle| \leq \varepsilon_\ell(\delta), \; \forall x,y \in \mathcal{A}_{\ell-1} \}$. By \cref{lemma 10}, we have
$$\mathbb{P}\left(E_{\ell}\right) \geq 1-\frac{\delta}{1+\lceil \log_2\log_2 T \rceil} \; .$$
By definition, $E = \cap_{\ell=1}^{B}E_{\ell}$, the following holds
$$\mathbb{P}(E^c) = \mathbb{P}(\overset{B}{\underset{\ell=1}{\cup}}E_{\ell}^{c}) \leq \sum_{\ell=1}^{B} \mathbb{P}(E_{\ell}^{c}) \leq \frac{B\delta}{1+\lceil \log_2\log_2 T \rceil} \leq \delta \; .$$
Therefore, we prove \cref{lemma:good-event}:
\[\mathbb{P}(E) \geq 1-\delta \; .\]
\end{proof}
\section{Proof of Lemma 3}
The proof of \cref{lemma:regularized-G-optimal-design} relies on the following four technical lemmas.
\begin{lemma}
\label{lemma 11} If $A : \mathbb{R} \rightarrow \mathbb{R}^{d\times d}$ is a $d\times d$ matrix, then the following holds
\[
\frac{d}{ds}\det(A(s)) = \text{trace}\left(\text{adj}(A)\frac{d}{ds}A(s)\right) \; .
\]
\end{lemma}
\begin{proof} Recall that $\det(A)$ can be expressed as a polynomial in the entries $a_{ij}$ of $A$. Concretely,
\[
\det(A) = \sum_{\sigma \in S_d} \text{sgn}(\sigma)a_{1,\sigma(1)}a_{2,\sigma(2)}\cdots a_{d,\sigma(d)}
\]
where the sum is over the symmetric group $S_d$ and $\text{sgn}(\sigma)$ denotes the signature of $\sigma$. Consequently, the partial derivative of $\det(A)$ with respect to a particular entry $a_{ij}$ involves only those terms in the polynomial that contain $a_{ij}$. From the definition of the adjugate, the $(j,i)$-th entry of $\text{adj}(A)$ is precisely the cofactor corresponding to $a_{ij}$. Concretely,
\[
\frac{\partial \det(A)}{\partial a_{ij}}
=
\bigl(\mathrm{adj}(A)\bigr)_{ji} \; . \tag{7}
\]
Next, let us regard $A$ as dependent on a variable $s$, so that each entry $a_{ij}$ becomes $a_{ij}(s)$. By the chain rule, we have
\[
\frac{d}{ds}\det(A(s)) = \sum_{1 \leq i,j \leq d} \frac{\partial \det(A)}{\partial a_{ij}} \cdot \frac{d}{ds}a_{ij}(s) \; .
\]
Substituting (7) into this sum yields
\[
\frac{d}{ds}\det(A(s))
=
\sum_{1\leq i,j \leq d} \bigl(\text{adj}(A)\bigr)_{ji}\cdot
\bigl(\tfrac{d}{ds}A(s)\bigr)_{ij} \; .
\]
Notice that the above expression is precisely the trace of the product \(\text{adj}(A) \cdot \tfrac{d}{ds}A(s)\). Indeed, writing matrix products in terms of their indices shows that
\[
\sum_{1\leq i,j \leq d} \bigl(\text{adj}(A)\bigr)_{ji} \cdot
\bigl(\tfrac{d}{ds}A(s)\bigr)_{ij}
=
\text{trace}\left(\text{adj}(A)\frac{d}{ds}A(s)\right) \; .
\]
Therefore, we prove \cref{lemma 11}:
\[
\frac{d}{ds}\det(A(s))
=
\text{trace}\left(\text{adj}(A)\frac{d}{ds}A(s)\right) \; .
\]
\end{proof}
\begin{lemma}
\label{lemma 12} Let $f:\mathcal{D}_\ell \rightarrow \mathbb{R}$ be defined as
\[
f(\pi_\ell) \coloneqq \log(\det(V_\ell(\pi_\ell))) = \log\left(\det\Bigg(\frac{\lambda}{c_\ell}I+\sum_{i=1}^{|\mathcal{A}_{\ell-1}|}\pi_{\ell i}T^{\frac{2^\ell-1}{2^\ell}}x^{(i)}{x^{(i)}}^T\Bigg)\right)\; .
\]
Then $f$ is a concave function.
\end{lemma}
\begin{proof} Define $\mathcal{T} \coloneqq \{X \in \mathbb{R}^{d\times d}:X > 0\}$, and let $g : \mathcal{T} \rightarrow \mathbb{R}$ be given by
\[
g(X) = \log(\det(X)) \; .
\]
The gradient of $g(X)$, using (7) and symmetry of $X$, is computed as
\[
\nabla_Xg(X) = \frac{\text{adj}(X)^T}{\det(X)} = \frac{\text{adj}(X)}{\det(X)} = X^{-1} \; .
\]
Using \cref{lemma 11}, for any symmetric matrix $U \in \mathbb{R}^{d\times d}$, the Hessian of $g(X)$ is computed as follows
\begin{align*}
\nabla^2_Xg(X)[U, U] &= \frac{d^2}{ds^2}\Big|_{s=0} \log(\det(X+sU)) \\
&= \frac{d}{ds}\Big|_{s=0} \frac{\text{trace}(\text{adj}(X+sU)U)}{\det(X+sU)} \\
&= \frac{d}{ds}\Big|_{s=0} \text{trace}((X+sU)^{-1}U) \\
&= -\text{trace}((X+sU)^{-1}U(X+sU)^{-1}U) \Big|_{s=0} \\
&= -\text{trace}(X^{-1}UX^{-1}U)
\end{align*}
Since $X^{-1}$ is positive definite and $U$ is symmetric, we have
\[
\text{trace}(X^{-1}UX^{-1}U) = \text{trace}(X^{-\frac{1}{2}}UX^{-\frac{1}{2}}X^{-\frac{1}{2}}UX^{-\frac{1}{2}}) = \text{trace}((X^{-\frac{1}{2}}UX^{-\frac{1}{2}})^TX^{-\frac{1}{2}}UX^{-\frac{1}{2}}) \geq 0 \; .
\]
Thus, the Hessian is negative semidefinite, which implies that $g(X)$ is concave for $X > 0$.
\\
Next, observe that \(V_\ell : \mathcal{D}_\ell \rightarrow \mathcal{T}\) is an affine map. Since the composition of a concave function with an affine map is concave, it follows that $f(\pi_\ell) = g(V_\ell(\pi_\ell))$ is concave. Specifically, for any
\(\pi_\ell,\tilde{\pi}_\ell \in \mathcal{D}_\ell\) and \(\lambda \in [0,1]\), we have
\[
g(V_\ell(\lambda\pi_\ell + (1-\lambda)\tilde{\pi}_\ell)) = g(\lambda V_\ell(\pi_\ell) + (1-\lambda) V_\ell(\tilde{\pi}_\ell)) \geq \lambda g(V_\ell(\pi_\ell)) + (1-\lambda)g(V_\ell(\tilde{\pi}_\ell)) \; .
\]
Therefore, $f(\pi_\ell)$ is a concave function of $\pi_\ell$. 
\end{proof}
\begin{lemma}
\label{lemma 13} Let $f:\mathbb{R}^d \rightarrow \mathbb{R}$ be a concave and differentiable function, and let $C \subseteq \mathbb{R}^d$ be a non-empty convex set. If \(x^* \in \text{argmax}_{x \in C} \, f(x)\), then the following holds
$$\langle x-x^*, \nabla f(x^*) \rangle \leq 0, \quad \forall x \in C \; .$$
\end{lemma}
\begin{proof} Define \(x(s) \coloneqq sx+(1-s)x^*\) for $s \in [0,1]$. Since $C$ is convex, we have $x(s) \in C$ for all $s \in [0,1]$. Let $\phi(s) \coloneqq f(x(s))$ for $s \in [0,1]$. Because $x^*$ is the maximizer of $f(x)$ over $C$, it follows that
$$\phi(0) = f(x^*) \geq f(x(s)) = \phi(s), \quad \forall s \in [0,1] \; .$$
Thus, $\phi(s)$'s right derivative at $s=0$ must be non-positive
\[0 \geq \phi'(0+) = \nabla f(x^*)^T (x-x^*) \; .\] 
Therefore, we prove \cref{lemma 13}:
$$\langle x-x^*, \nabla f(x^*) \rangle \leq 0, \quad \forall x \in C \; .$$
\end{proof}
\begin{lemma}
\label{lemma 14} After the $\ell$-th batch $(\ell \geq 1)$, the following bound holds
\[\underset{x \in \mathcal{A}_{\ell-1}}{\max} \|x\|^2_{V_{\ell}(\pi_\ell^*)^{-1}} \leq \frac{dc_\ell}{d\lambda + T^{\frac{2^\ell-1}{2^\ell}}c_\ell} \; .\]
\end{lemma}
\begin{proof} Let us define $f(\pi_\ell) \coloneqq \log(\det(V_\ell(\pi_\ell))$. We compute the gradient of $f(\pi_\ell)$ component-wise
\begin{align*}
(\nabla f(\pi_\ell))_i &= \frac{\partial}{\partial \pi_{\ell i}}\log(\det(V_\ell(\pi_\ell)) \\
&= \frac{1}{\det(V_\ell(\pi_\ell))} \cdot \frac{\partial}{\partial \pi_{\ell i}}\det(V_\ell(\pi_\ell)) \; .
\end{align*}
Using \cref{lemma 11}, we obtain
\begin{align*}
(\nabla f(\pi_\ell))_i &= \frac{1}{\det(V_\ell(\pi_\ell))} \cdot \text{trace}\left(\text{adj}(V_\ell(\pi_\ell))\frac{\partial V_\ell(\pi_\ell)}{\partial \pi_{\ell i}}\right)\\
&= \text{trace}\left(V_\ell(\pi_\ell)^{-1} \cdot \frac{\partial V_\ell(\pi_\ell)}{\partial \pi_{\ell i}}\right)\\
&= \text{trace}(V_\ell(\pi_\ell)^{-1} \cdot T^{\frac{2^\ell-1}{2^\ell}} x^{(i)}{x^{(i)}}^T) \\
&= T^{\frac{2^\ell-1}{2^\ell}}\|x^{(i)}\|_{V_\ell(\pi_\ell)^{-1}}^2 \; .
\end{align*}
Since $V_\ell(\pi_\ell)$ is a linear function of $\pi_\ell$, the function $f(\pi_\ell)$ is continuous. Furthermore, as the domain of $f$ is the compact set $\mathcal{D}_\ell$, the Weierstrass theorem guarantees the existence of a maximizer $\pi_\ell' \in \text{argmax}_{\pi_\ell \in \mathcal{D}_\ell} f(\pi_\ell)$. Additionally, it is well known that matrix inversion is a continuous operation on the set of invertible matrices, which implies that $V_\ell(\pi_\ell)^{-1}$ is a continuous function of $\pi_\ell$. Since the maximum of finitely many continuous functions is also continuous and the set $\mathcal{D}_\ell$ is compact, the Weierstrass theorem ensures the existence of a minimizer $\pi_\ell^* \in \text{argmin}_{\pi_\ell \in \mathcal{D}_\ell}\max_{x\in \mathcal{A}_{\ell-1}}\|x\|_{V_\ell(\pi_\ell)^{-1}}^2$.
\\
Since $\mathcal{D}_\ell$ is compact and $f(\pi_\ell)$ is concave by \cref{lemma 12}, \cref{lemma 13} implies
\[
0 \geq \langle \nabla f(\pi_\ell'), \pi_\ell-\pi_\ell' \rangle, \quad \forall \pi_\ell \in \mathcal{D}_\ell \; .
\]
Expanding this inequality yields
\[
0 \geq \sum_{i=1}^{|\mathcal{A}_{\ell-1}|} (\nabla f(\pi_\ell'))_i (\pi_{\ell i} - \pi_{\ell i}') 
= \sum_{i=1}^{|\mathcal{A}_{\ell-1}|} T^{\frac{2^\ell-1}{2^\ell}}\|x^{(i)}\|_{V_\ell(\pi_\ell')^{-1}}^2 (\pi_{\ell i} - \pi_{\ell i}') , \quad \forall \pi_\ell \in \mathcal{D}_\ell \; .
\]
When we set $\pi_\ell$ to be a Dirac distribution $(0,\dots,1,\dots,0)$ and divide by $T^{\frac{2^\ell-1}{2^\ell}}$, we obtain
\begin{align*}
0 &\geq \|x^{(i)}\|_{V_\ell(\pi_\ell')^{-1}}^2 - \sum_{j=1}^{|\mathcal{A}_{\ell-1}|} \|x^{(j)}\|_{V_\ell(\pi_\ell')^{-1}}^2 \pi_{\ell j}' \\
&= \|x^{(i)}\|_{V_\ell(\pi_\ell')^{-1}}^2 - \sum_{j=1}^{|\mathcal{A}_{\ell-1}|} \text{trace}(\pi_{\ell j}'x^{(j)}{x^{(j)}}^T V_\ell(\pi_\ell')^{-1}) \\
&= \|x^{(i)}\|_{V_\ell(\pi_\ell')^{-1}}^2 - T^{-\frac{2^\ell-1}{2^\ell}}\text{trace}(\sum_{j=1}^{|\mathcal{A}_{\ell-1}|} \pi_{\ell j}'T^{\frac{2^\ell-1}{2^\ell}}x^{(j)}{x^{(j)}}^T V_\ell(\pi_\ell')^{-1}) \\
&= \|x^{(i)}\|_{V_\ell(\pi_\ell')^{-1}}^2 - T^{-\frac{2^\ell-1}{2^\ell}}\text{trace}((V_\ell(\pi_\ell') - \frac{\lambda}{c_\ell} I )V_\ell(\pi_\ell')^{-1}) \\
&= \|x^{(i)}\|_{V_\ell(\pi_\ell')^{-1}}^2 - T^{-\frac{2^\ell-1}{2^\ell}}\text{trace}(I - \frac{\lambda}{c_\ell} V_\ell(\pi_\ell')^{-1}) \\
&= \|x^{(i)}\|_{V_\ell(\pi_\ell')^{-1}}^2 - dT^{-\frac{2^\ell-1}{2^\ell}} + \frac{\lambda T^{-\frac{2^\ell-1}{2^\ell}}}{c_\ell}\text{trace}(V_\ell(\pi_\ell')^{-1}) \; . \tag{8}
\end{align*}
Let $\mathcal{\lambda}_1',\mathcal{\lambda}_2',\cdots,\mathcal{\lambda}_d'$ be the eigenvalues of $V_\ell(\pi_\ell')$. By the Cauchy-Schwarz inequality, we have
\[
\text{trace}(V_\ell(\pi_\ell')^{-1}) = \sum_{i=1}^{d}\frac{1}{\lambda_i'} \geq \frac{d^2}{\sum_{i=1}^{d} \lambda_i'} = \frac{d^2}{\text{trace}(V_\ell(\pi_\ell'))} \; .
\]
Using Assumption~\ref{assum:norm}, we obtain
\begin{align*}
\text{trace}(V_\ell(\pi_\ell')) &= \text{trace}\left(\frac{\lambda}{c_\ell}I+\sum_{i=1}^{|\mathcal{A}_{\ell-1}|}\pi_{\ell i}'T^{\frac{2^\ell-1}{2^\ell}}x^{(i)}{x^{(i)}}^T\right) \\
&= \frac{d\lambda}{c_\ell}+\sum_{i=1}^{|\mathcal{A}_{\ell-1}|}\pi_{\ell i}'T^{\frac{2^\ell-1}{2^\ell}}\|x^{(i)}\|^2 \\
&\leq \frac{d\lambda}{c_\ell} + T^{\frac{2^\ell-1}{2^\ell}} \; ,
\end{align*}
which implies 
\[
\text{trace}(V_\ell(\pi_\ell')^{-1}) \geq \frac{d^2 c_\ell}{d\lambda + T^{\frac{2^\ell-1}{2^\ell}}c_\ell} \; . \tag{9}
\]
Combining (8) and (9), we obtain
\[
\|x^{(i)}\|_{V_\ell(\pi_\ell')^{-1}}^2 \leq dT^{-\frac{2^\ell-1}{2^\ell}} -\frac{d^2\lambda T^{-\frac{2^\ell-1}{2^\ell}}}{d\lambda + T^{\frac{2^\ell-1}{2^\ell}}c_\ell} = \frac{dc_\ell}{d\lambda + T^{\frac{2^\ell-1}{2^\ell}}c_\ell} \; .
\]
\\
Since this inequality holds for all $1 \leq i \leq |\mathcal{A}_{\ell-1}|$ and $\pi_\ell^*$ minimizes $\max_{x \in \mathcal{A}_{\ell-1}} \|x\|_{V_\ell(\pi_\ell)^{-1}}^2$ over $\mathcal{D}_\ell$, where
\[
\pi_\ell^* \in \underset{\pi_\ell \in \mathcal{D}_\ell}{\text{argmin}} \underset{x \in \mathcal{A}_{\ell-1}}{\text{max}} \|x\|_{V_\ell(\pi_\ell)^{-1}}^2 \; ,
\]
we conclude \cref{lemma 14}:
\[
\underset{x \in \mathcal{A}_{\ell-1}}{\max} \|x\|^2_{V_{\ell}(\pi_\ell^*)^{-1}}\leq \underset{x \in \mathcal{A}_{\ell-1}}{\max} \|x\|^2_{V_{\ell}(\pi_\ell')^{-1}} \leq \frac{dc_\ell}{d\lambda + T^{\frac{2^\ell-1}{2^\ell}}c_\ell} \; .
\]
\end{proof}
\begin{proof}[Proof of Lemma 3.] Define $\alpha_i \coloneqq \lceil T^{\frac{2^\ell-1}{2^\ell}} (c_\ell \pi_{\ell i}^* + (1-c_\ell) \cdot \bbone_{\{1\}}(i)) \rceil - c_\ell T^{\frac{2^\ell-1}{2^\ell}} \pi_{\ell i}^*$ for $1 \leq i \leq |\mathcal{A}_{\ell-1}|$, and let $X \coloneqq [\sqrt{\alpha_1}x^{(1)}, \sqrt{\alpha_2}x^{(2)}, \dots,\sqrt{\alpha_{|\mathcal{A}_{\ell-1}|}}x^{(|\mathcal{A}_{\ell-1}|)} ]$ be a matrix whose columns are the vectors $\sqrt{\alpha_i}x^{(i)}$. Then, the following holds
$$H_\ell = \lambda I + \sum_{i=1}^{|\mathcal{A}_{\ell-1}|} \lceil T^{\frac{2^\ell-1}{2^\ell}}(c_\ell \pi_{\ell i}^* + (1-c_\ell)\cdot \bbone_{\{1\}}(i)) \rceil x^{(i)}{x^{(i)}}^T = c_\ell V_\ell(\pi_{\ell}^*) + \sum_{i=1}^{|\mathcal{A}_{\ell-1}|} \alpha_i x^{(i)}{x^{(i)}}^T = c_\ell V_\ell(\pi_{\ell}^*) + XX^T \; .$$
\\
Using the Sherman-Morrison-Woodbury formula and the positive definiteness of $(I+\frac{1}{c_\ell}X^T V_{\ell}(\pi_\ell^*)^{-1} X)^{-1}$, we have
$$H_{\ell}^{-1} = \frac{1}{c_\ell}V_{\ell}(\pi_\ell^*)^{-1} - \frac{1}{c_\ell^2}V_{\ell}(\pi_\ell^*)^{-1}X(I+\frac{1}{c_\ell}X^T V_{\ell}(\pi_\ell^*)^{-1} X)^{-1}X^TV_{\ell}(\pi_\ell^*)^{-1} \leq \frac{1}{c_\ell}V_{\ell}(\pi_\ell^*)^{-1} \; .$$
Therefore, we prove \cref{lemma:regularized-G-optimal-design}:
\[ \underset{x\in \mathcal{A}_{\ell-1}}{\max}\|x\|_{H_{\ell}^{-1}}^2 \leq \underset{x\in \mathcal{A}_{\ell-1}}{\max}\|x\|_{\frac{1}{c_\ell}V_{\ell}(\pi_{\ell}^*)^{-1}}^2 \leq \frac{d}{d\lambda + T^{\frac{2^\ell-1}{2^\ell}}c_\ell} \; .
\]
\end{proof}
\section{Proof of Theorem 1}
\begin{proof}[Proof of Theorem 1] Define the nice event $E$ as 
\[E \coloneqq \bigcap\limits_{\ell = 1}^{B}\left\{|\langle x-y,\hat{\theta}_\ell - \theta^* \rangle| \leq \varepsilon_\ell, \; \forall x,y \in \mathcal{A}_{\ell-1}\right\} \; .
\] 
By \cref{lemma:good-event}, this nice event occurs with high probability
$$\mathbb{P}(E) \geq 1-\frac{1}{T} \; .$$
We can decompose the cumulative expected regret using the nice event $E$ as follows
\[
\mathcal{R}(T) = \mathbb{E}\left[\sum_{t=1}^T\langle x^*-x_t, \theta^* \rangle\bigg| E^c \right] \cdot \mathbb{P}(E^c) + \mathbb{E}\left[\sum_{t=1}^T\langle x^*-x_t, \theta^* \rangle\bigg| E \right] \cdot \mathbb{P}(E) \; .
\]
Using Assumption~\ref{assum:norm}, and the Cauchy-Schwarz inequality, we obtain
\begin{align*}
\langle x^*-x_t, \theta^* \rangle &\leq |\langle x^*-x_t, \theta^* \rangle| \\
&\leq |\langle x^*, \theta^* \rangle| + |\langle x_t, \theta^* \rangle| \\
&\leq \|x^*\|\cdot\|\theta^*\| + \|x_t\|\cdot\|\theta^*\| \\
&\leq 2 \; \tag{10}.
\end{align*}
Therefore, the following bound on cumulative expected regret holds
\begin{align*}
\mathcal{R}(T) &\leq 2T\cdot \mathbb{P}(E^c)+\mathbb{E}\left[\sum_{t=1}^T\langle x^*-x_t, \theta^* \rangle\bigg| E \right] \cdot \mathbb{P}(E) \\
&\leq \Ocal(1) + \mathbb{E}\left[\sum_{t=1}^T\langle x^*-x_t, \theta^* \rangle\bigg| E \right] \; .
\end{align*}
From now on, we assume that $E$ occurs. We claim that $x^* \in \mathcal{A}_{\ell}$ for $1 \leq \ell \leq B$. We prove this by mathematical induction. For $\ell = 1$, since $x^* \in \mathcal{A}_0$, we obtain
\begin{align*}
\langle \hat{\theta}_{1}, \hat{x}_{1}^*-x^* \rangle
&= \langle \hat{\theta}_{1} - \theta^* + \theta^*, \hat{x}_{1}^*-x^* \rangle \\
&= \langle \hat{\theta}_{1} - \theta^*, \hat{x}_{1}^* - x^* \rangle + \langle \theta^*, \hat{x}_{1}^* - x^* \rangle \\
&\leq \varepsilon_1 \; . 
\end{align*}
Therefore, $x^* \in \mathcal{A}_1$ holds, establishing the base case. Now assume $x^* \in \mathcal{A}_{\ell-1}$. Then
\begin{align*}
\langle \hat{\theta}_{\ell}, \hat{x}_{\ell}^*-x^* \rangle
&= \langle \hat{\theta}_{\ell} - \theta^* + \theta^*, \hat{x}_{\ell}^*-x^* \rangle \\
&= \langle \hat{\theta}_{\ell} - \theta^*, \hat{x}_{\ell}^* - x^* \rangle + \langle \theta^*, \hat{x}_{\ell}^* - x^* \rangle \\
&\leq \varepsilon_\ell \; .
\end{align*}
Therefore, $x^* \in \mathcal{A}_{\ell}$ holds. By induction, we conclude that $x^* \in \mathcal{A}_{\ell}$ for all $1 \leq \ell \leq B$.
\\
For each suboptimal arm $x(\neq x^*)$, define $\ell_x \coloneqq \min\{\ell\,|\, \Delta_x > 2\varepsilon_\ell\}$ as the first phase where the suboptimality gap of arm $x$ is larger than $2\varepsilon_\ell$. If a suboptimal arm $x(\neq x^*)$ is not eliminated in the first $\ell_{x}-1$ batches, after the $\ell_x$-th batch, we obtain
\begin{align*}
\langle \hat{\theta}_{\ell_x}, \hat{x}_{\ell_x}^{*}- \; x \rangle &\geq \langle \hat{\theta}_{\ell_x}, x^*- \;x \rangle \\
&= \langle \hat{\theta}_{\ell_x}-\; \theta^*+\theta^*, x^*-\; x \rangle \\
&= \langle \hat{\theta}_{\ell_x}-\; \theta^*, x^*-\;x \rangle + \langle \theta^*, x^* - \;x \rangle \\
&\geq -\varepsilon_{\ell_x} + \Delta_x \\
&> \varepsilon_{\ell_x} \; .
\end{align*}
Therefore, we know suboptimal arm $x(\neq x^*)$ will be eliminated in the first $\ell_x$ batches
$$x \notin \mathcal{A}_{\ell_x}, \quad \forall x \in  \mathcal{A}_0 \backslash \{x^*\} \; .$$
Thus, for $\ell \geq 2$, the suboptimality gap of any arm included in $\mathcal{A}_{\ell-1}$ is smaller than $2\varepsilon_{\ell-1}$ in $\ell$-th batch, where
\[
\Delta_x \leq 2\varepsilon_{\ell-1}, \quad \forall x \in \mathcal{A}_{\ell-1} \; . \tag{11}
\]
Using the triangle inequality, we obtain the following bound for $\ell \geq 2$
\begin{align*}
M_\ell &= \underset{x,y \in \mathcal{A}_{\ell-1}}{\max}\|x-y\|_{H_\ell^{-1}}^2 \\
&\leq \underset{x,y \in\mathcal{A}_{\ell-1}}{\max}(\|x\|_{H_{\ell}^{-1}}+\|y\|_{H_{\ell}^{-1}})^2 \\
&\leq \underset{x \in\mathcal{A}_{\ell-1}}{\max} 4\|x\|_{H_{\ell}^{-1}}^2 \; . \tag{12}
\end{align*}
Let $\text{Regret}_{\ell}$ denote the cumulative expected regret of the $\ell$-th batch when $E$ occurs. Combining (11) and (12), we obtain the following bound of cumulative expected regret
\begin{align*}
\sum_{\ell=2}^{B}\text{Regret}_\ell &\leq \sum_{\ell=2}^{B} (T^{\frac{2^\ell-1}{2^{\ell}}}+|\mathcal{A}_{\ell-1}|) 2\varepsilon_{\ell-1} \\
&\leq \sum_{\ell=2}^{B} \frac{2\sqrt{M_{\ell-1}}\left(\sqrt{(d-1)\log(\frac{4}{4\zeta^2-\zeta^4})+2\log(2T\sqrt{2\pi d}(1+\lceil \log_2\log_2 T \rceil))} +\sqrt{\lambda}\right)}{(1-\zeta)/(T^{\frac{2^\ell-1}{2^{\ell}}}+|\mathcal{A}_{\ell-1}|)} \\
&\leq \sum_{\ell=2}^{B} \frac{4\underset{x\in \mathcal{A}_{\ell-2}}{\max}\|x\|_{H_{\ell-1}^{-1}}\left(\sqrt{(d-1)\log(\frac{4}{4\zeta^2-\zeta^4})+2\log(2T\sqrt{2\pi d}(1+\lceil \log_2\log_2 T \rceil))} +\sqrt{\lambda}\right)}{(1-\zeta)/(T^{\frac{2^\ell-1}{2^{\ell}}}+|\mathcal{A}_{\ell-1}|)} \\
&\leq \sum_{\ell=2}^{B} \frac{4\sqrt{d}(T^{\frac{2^\ell-1}{2^{\ell}}}+|\mathcal{A}_{\ell-1}|)\left(\sqrt{(d-1)\log(\frac{4}{4\zeta^2-\zeta^4})+2\log(2T\sqrt{2\pi d}(1+\lceil \log_2\log_2 T \rceil))} +\sqrt{\lambda}\right)}{(1-\zeta)\sqrt{d\lambda+T^{\frac{2^{\ell-1}-1}{2^{\ell-1}}}c_{\ell-1}}}
\; .
\end{align*}
Here, we use 
\[\varepsilon_\ell \leq \sqrt{M_\ell} \cdot \frac{\sqrt{(d-1)\log(\frac{4}{4\zeta^2-\zeta^4})+2\log(2T\sqrt{2\pi d}(1+\lceil \log_2\log_2 T \rceil))} +\sqrt{\lambda}}{1-\zeta} \; .\]
Using $T\geq \Omega(K^\frac{4}{5})$ and $B \leq 1+ \lceil \log_2\log_2 T \rceil$, we derive the following bound for the cumulative expected regret incurred during the batches with $\ell \geq 2$
\begin{align*}
\sum_{\ell=2}^{B}\text{Regret}_\ell &= \sum_{\ell=2}^{B} \Ocal\left(\frac{(T^{\frac{2^\ell-1}{2^\ell}} + K)\left(\sqrt{d^2 + d\log(T\sqrt{d}\log\log T)} + \sqrt{d}\right)}{\sqrt{d+T^{\frac{2^{\ell-1}-1}{2^{\ell-1}}}}}\right) \\
&= \sum_{\ell=2}^{B} \Ocal\left(\frac{(\sqrt{T}+KT^{-\frac{2^{\ell-1}-1}{2^\ell}})\left(\sqrt{d^2 + d\log(T\log\log T)} + \sqrt{d}\right)}{\sqrt{dT^{-\frac{2^{\ell-1}-1}{2^{\ell-1}}}+1}}\right) \\
&= \Ocal\left(\frac{(B-1)(\sqrt{T} + KT^{-\frac{3}{4}})\sqrt{d^2 + d\log T}}{\sqrt{dT^{-\frac{2^{B-1}-1}{2^{B-1}}}+1}}\right) \\
&= \Ocal\left(\frac{(\sqrt{dT} + \sqrt{d}KT^{-\frac{3}{4}})\sqrt{d + \log T}\log\log T}{\sqrt{dT^{-1}+1}}\right) \\
&= \Ocal\left(\sqrt{dT}\sqrt{d + \log T}\log\log T\right) \\
&= \tilde{\Ocal}(d\sqrt{T}) \; .
\end{align*}
We can derive a similar upper bound using an alternative bound for $\varepsilon_\ell$,
\[
\varepsilon_\ell \leq \sqrt{M_\ell} \cdot \left(\sqrt{2\log\left(T(|\mathcal{A}_{\ell-1}|^2-|\mathcal{A}_{\ell-1}|)(1+\lceil \log_2\log_2T \rceil)\right)} + \sqrt{\lambda} \right)\; .
\]
This leads to the following bound for the cumulative regret after the second batch
\begin{align*}
\sum_{\ell=2}^{B}\text{Regret}_\ell &\leq \sum_{\ell=2}^{B} (T^{\frac{2^\ell-1}{2^{\ell}}}+|\mathcal{A}_{\ell-1}|) 2\varepsilon_{\ell-1} \\
&\leq \sum_{\ell=2}^{B} {2\sqrt{M_{\ell-1}}}(T^{\frac{2^\ell-1}{2^{\ell}}}+|\mathcal{A}_{\ell-1}|) \left(\sqrt{2\log(T(|\mathcal{A}_{\ell-2}|^2-|\mathcal{A}_{\ell-2}|)(1+\lceil \log_2\log_2T \rceil))}+\sqrt{\lambda}\right) \\
&\leq \sum_{\ell=2}^{B} {4\underset{x\in \mathcal{A}_{\ell-2}}{\max}\|x\|_{H_{\ell-1}^{-1}}}(T^{\frac{2^\ell-1}{2^{\ell}}}+|\mathcal{A}_{\ell-1}|) \left(\sqrt{2\log(T|\mathcal{A}_{\ell-2}|^2(1+\lceil \log_2\log_2T \rceil))}+\sqrt{\lambda}\right) \\
&\leq \sum_{\ell=2}^{B} \frac{4\sqrt{d}(T^{\frac{2^\ell-1}{2^{\ell}}}+|\mathcal{A}_{\ell-1}|)\left(\sqrt{2\log(T|\mathcal{A}_{\ell-2}|^2(1+\lceil \log_2\log_2T \rceil))}+\sqrt{\lambda}\right)}{\sqrt{d\lambda+T^{\frac{2^{\ell-1}-1}{2^{\ell-1}}}c_{\ell-1}}}
\end{align*}
Using $T\geq \Omega(K^\frac{4}{5})$ and $B \leq 1+ \lceil \log_2\log_2 T \rceil$, we derive the following bound for the cumulative expected regret incurred during the batches with $\ell \geq 2$
\begin{align*}
\sum_{\ell=2}^{B}\text{Regret}_\ell &= \sum_{\ell=2}^{B} \Ocal\left(\frac{\sqrt{d}(T^{\frac{2^\ell-1}{2^\ell}}+K)\left(\sqrt{\log(KT\log\log T)} +1\right)}{\sqrt{d+T^{\frac{2^{\ell-1}-1}{2^{\ell-1}}}}}\right) \\
&= \sum_{\ell=2}^{B} \Ocal\left(\frac{(\sqrt{dT} + \sqrt{d}KT^{-\frac{2^{\ell-1}-1}{2^{\ell}}})\left(\sqrt{\log(KT\log\log T)} +1\right)}{\sqrt{dT^{-\frac{2^{\ell-1}-1}{2^{\ell-1}}}+1}}\right) \\
&= \Ocal\left(\frac{(B-1)(\sqrt{dT} + \sqrt{d}KT^{-\frac{3}{4}})\left(\sqrt{\log(KT\log\log T)} +1\right)}{\sqrt{dT^{-\frac{2^{B-1}-1}{2^{B-1}}}+1}}\right) \\
&= \Ocal\left(\frac{(\sqrt{dT} + \sqrt{d}KT^{-\frac{3}{4}})\sqrt{\log(KT)}\log\log T}{\sqrt{dT^{-1}+1}}\right) \\
&= \Ocal\left({\sqrt{dT\log(KT)}\log\log T}\right) \\
&= \tilde{\Ocal}(\sqrt{dT\log K}) \; . \\
\end{align*}
Thus, the regret except for the first batch is bounded by
$$\sum_{\ell=2}^{B}\text{Regret}_\ell = \Ocal\big( \sqrt{dT}\big(\sqrt{\log(KT)}\wedge \sqrt{d+\log T}\big)\log\log T \big) = \tilde{\Ocal}\big(\sqrt{dT\log K}\wedge d\sqrt{T}\big) \; .$$
Using (10) and $T \geq \Omega(\frac{K^2}{(d \, \wedge \, \log K)d})$, the cumulative expected regret for the first batch can be easily bounded as follows
$$\text{Regret}_1 \leq 2(\sqrt{T}+K) = \Ocal\big(\sqrt{dT\log K}\wedge d\sqrt{T}\big) \; .$$
Thus, the worst-case regret is given by 
\[
\mathcal{R}(T) = \Ocal\big( \sqrt{dT}\big(\sqrt{\log(KT)}\wedge \sqrt{d+\log T}\big)\log\log T \big)  = \tilde{\Ocal}\big(\sqrt{dT\log K}\wedge d\sqrt{T}\big) \; .
\] 
\end{proof}

\section{Additional Experiments}\label{sec:additional-experiments}
\subsection{Runtime Comparison Analysis} \label{sec:runtime-comparison}
We provide a runtime comparison with batched linear bandit algorithms in \citet{ruan2021linear,hanna2023contexts,zhang2025almost}. The experiments were conducted with $K=50$ arms and a feature dimension of $d=5$, where all arms and the unknown parameter $\theta^*$ were sampled from a normal distribution. Under the same experimental setting, the \texttt{BLAE} algorithm requires an average of $0.73$ seconds over 10 runs when $T =$ 100,000. 
In contrast, as shown in Table~\ref{table:runtime-with-Ruan-and-Hanna}, the algorithm in \citet{ruan2021linear} requires an average of $464.2$ seconds for $T=5000$, while the algorithm in \citet{zhang2025almost} takes an average of $717.8$ seconds for $T=5000$, demonstrating their limitation in large-scale problems. 
Furthermore, the algorithm in \citet{hanna2023contexts} faces severe computational challenges due to the construction of $\Theta'$ as mentioned in Section~\ref{sec:reason of fail}. The size of $\Theta'$ is $\mathcal{\Ocal}(T^d)$, meaning that for $T=1000$, $\Theta'$ would contain approximately $10^{15}$ vectors, making the algorithm infeasible even for moderate values of $T$. 
Due to these computational constraints, we were unable to measure its runtime. From this experiment, we conclude that the batched linear bandit algorithms proposed in \citet{ruan2021linear,zhang2025almost,hanna2023contexts} are impractical for our setting.
\begin{table}[!htbp]  
\caption{Average runtime (seconds) over 10 runs for the normal distribution setting with $K=50$ and $d=5$.}
\label{table:runtime-with-Ruan-and-Hanna}
\begin{center}
\begin{small}
\begin{tabular}{lrrrr}
\toprule
& \multicolumn{4}{c}{\textbf{Runtime}} \\
\textbf{Time step}
& \citet{ruan2021linear} & \citet{hanna2023contexts} & \citet{zhang2025almost} & BLAE (Ours)\\
\midrule
$T=500$    & $89.2$ & $\geq$ 10000  & $57.1$ & $0.26$  \\
$T=1000$    & $132.4$ & $\geq$ 10000 &  $88.9$ & $0.35$ \\
$T=5000$   & $464.2$ & $\geq$ 10000 & $717.8$ & $0.55$   \\
\bottomrule
\end{tabular}
\end{small}
\end{center}
\vskip -0.1in
\end{table}

\subsection{Additional Comparisons of  Regret and Batch Complexity}

In this section, we provide additional experimental results for the instances where arms are randomly sampled from a Gaussian distribution, following the same format as in Figure~\ref{uniform figure}. Similar to the uniform distribution case, we repeat each experiment 10 times, independently resampling the arms from the Gaussian distribution in every run. We observe similar trends under the Gaussian distribution setting. As shown in Figure~\ref{normal figure}, the gap in mean regret is even more pronounced compared to the uniform distribution case, further demonstrating the practical superiority of the \texttt{BLAE} algorithm. It consistently outperforms all other algorithms, achieving the lowest mean regret and the highest stability, even in comparison to \texttt{RS-OFUL}, which utilizes over 100 batches. In contrast, the algorithm proposed in \citet{ren2024optimal} demonstrates the weakest empirical performance and notable instability over 80 independent trials, particularly resembling the behavior observed under uniform distribution case.
\begin{figure*}[!htbp]
\centering
\begin{subfigure}{\textwidth}
\includegraphics[width=\textwidth]{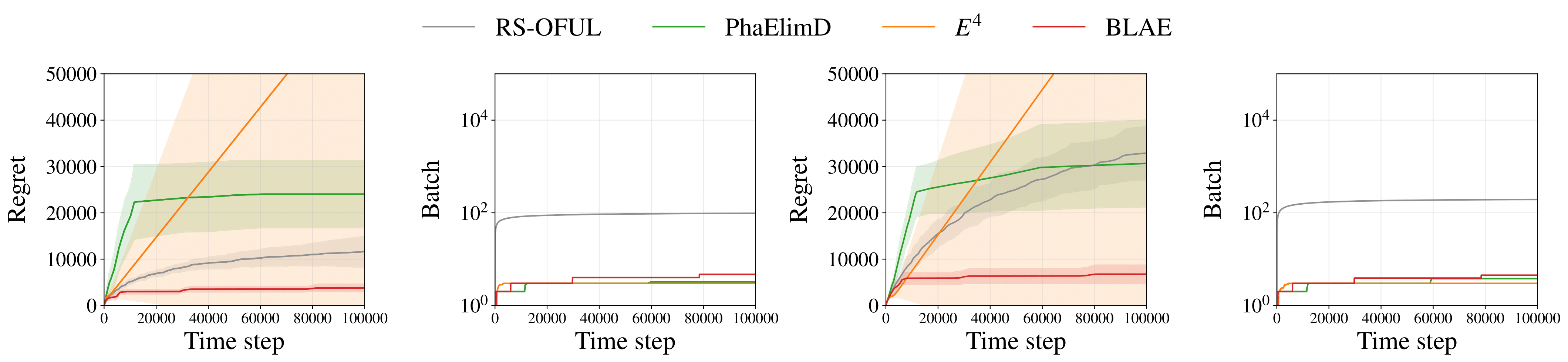}
    \begin{center}
        \makebox[0.25\textwidth][c]{(a) $K = 50, d=5$} 
        \makebox[0.25\textwidth][c]{}
        \makebox[0.25\textwidth][c]{(b) $K = 50, d=10$}
    \end{center}
\end{subfigure}
\begin{subfigure}{\textwidth}
\includegraphics[width=\textwidth]{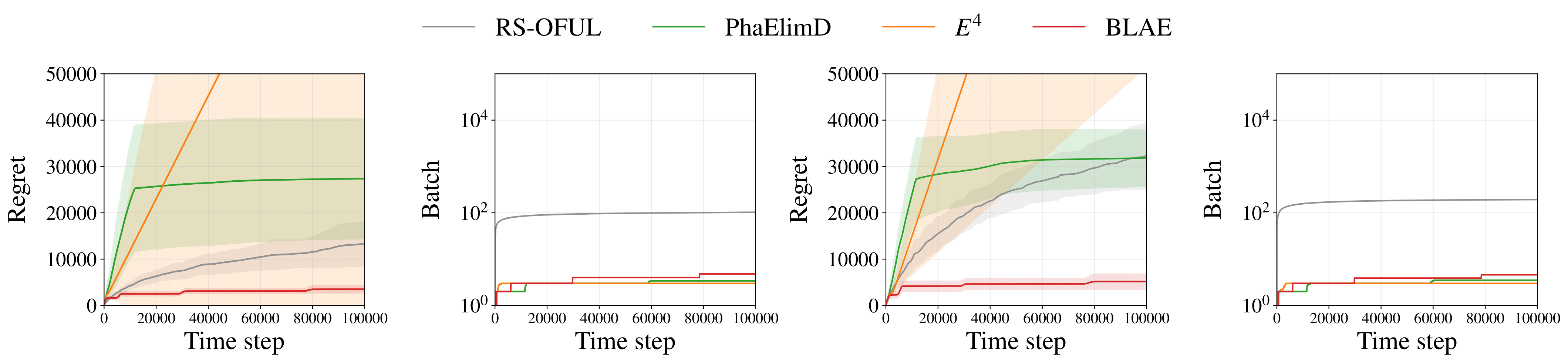}
    \begin{center}
        \makebox[0.25\textwidth][c]{(c) $K = 100, d=5$} 
        \makebox[0.25\textwidth][c]{}
        \makebox[0.25\textwidth][c]{(d) $K = 100, d=10$}
    \end{center}
\end{subfigure}
\end{figure*}
\begin{figure*}[!htbp]
\begin{subfigure}{\textwidth}
\includegraphics[width=\textwidth]{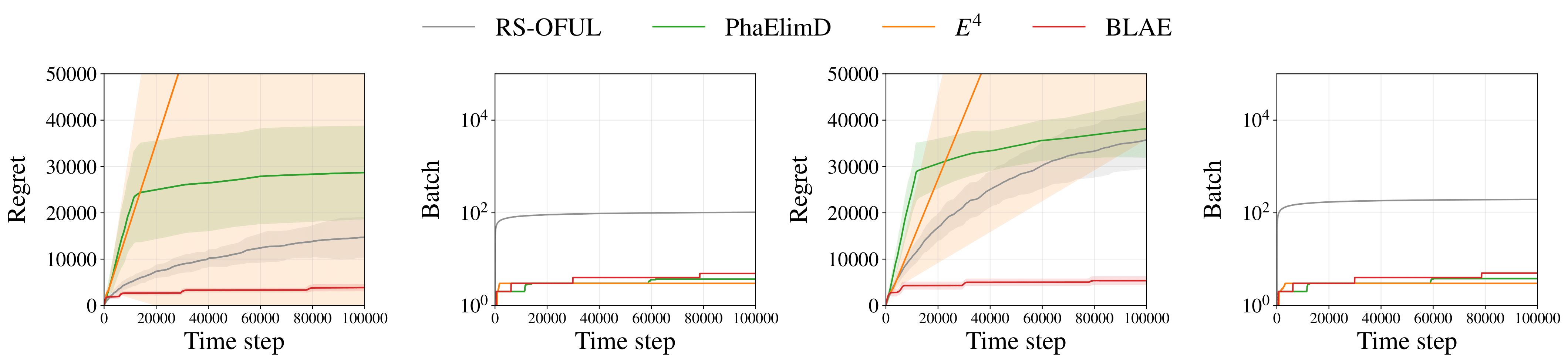}
    \begin{center}
        \makebox[0.25\textwidth][c]{(c) $K = 200, d=5$} 
        \makebox[0.25\textwidth][c]{}
        \makebox[0.25\textwidth][c]{(d) $K = 200, d=10$}
    \end{center}
\end{subfigure}
\begin{subfigure}{\textwidth}
\includegraphics[width=\textwidth]{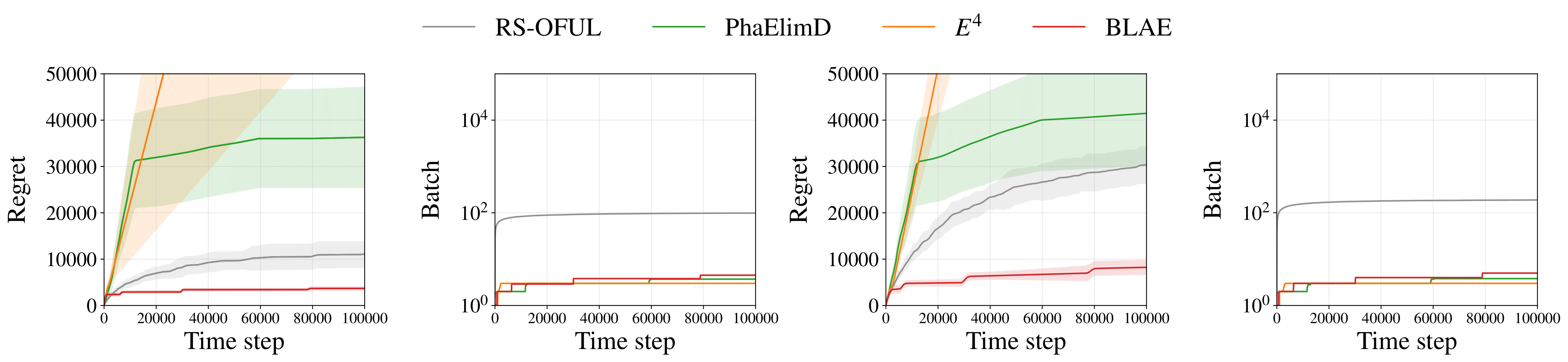}
    \begin{center}
        \makebox[0.25\textwidth][c]{(c) $K = 400, d=5$} 
        \makebox[0.25\textwidth][c]{}
        \makebox[0.25\textwidth][c]{(d) $K = 400, d=10$}
    \end{center}
\end{subfigure}
\setcounter{figure}{1}
\captionof{figure}{Regret and batch complexity over time for different values of $K$ and $d$}
\label{normal figure}
\end{figure*}
\end{document}